\documentclass[journal]{IEEEtran}
\usepackage{amsmath}
\usepackage{amsthm}
\usepackage{amssymb}
\usepackage{cite}
\usepackage{graphicx}
\usepackage{graphics}
\usepackage{epstopdf}
\usepackage{color}
\usepackage{soul}
\usepackage{xcolor}
\usepackage{bm}
\usepackage{hhline}
\usepackage{balance}
\usepackage{multirow}
\usepackage{booktabs}
\usepackage{subfigure}
\usepackage{url}
\usepackage[ruled]{algorithm2e}
\usepackage[colorlinks,linkcolor=blue]{hyperref}
\usepackage{silence}
\usepackage{bbding}
\usepackage{tasks}
\usepackage{footnote}
\usepackage{fancyhdr}
\settasks{
  style=itemize,
  after-item-skip=0pt
}
\newtheoremstyle{mythm}{1.5ex plus 1ex minus .2ex}{1.5ex plus 1ex minus .2ex}{}{}{\itshape}{:}{5pt plus 1pt minus 1pt}{}
\theoremstyle{mythm}

\allowdisplaybreaks

\graphicspath{{image/}}

\begin{document}

\title{AutoGMap: Learning to Map Large-scale Sparse Graphs on Memristive Crossbars}
	\author{Bo Lyu, Shengbo Wang, Shiping Wen,~\IEEEmembership{Senior Member,~IEEE}, Kaibo Shi, Yin Yang, Lingfang Zeng\\ and Tingwen Huang,~\IEEEmembership{Fellow,~IEEE} 	
	
	\thanks{This publication was made possible by NPRP grant: NPRP 9-466-1-103 from Qatar National Research Fund. The statements made herein are solely the responsibility of the authors. \textit{(Corresponding authors: Shiping Wen.)}
}
	
		\thanks{B. Lyu and L. Zeng are with Zhejiang Lab, Hangzhou, China, 311121, (email: \{bo.lyu; zenglf\}@zhejianglab.com). S. Wang is with School of Computer Science and Engineering, University of Electronic Science and Technology of China, Chengdu 611731, China (e-mail: shnbo.wang@foxmail.com). S. Wen is with Australian AI Institute, Faculty of Engineering and Information Technology, University of Technology Sydney, NSW 2007, Australia (shiping.wen@uts.edu.au). K. Shi is with School of Information Science and Engineering, Chengdu University, Chengdu, 611040, China, (email: skbs111@163.com). 
		Y. Yang is with College of Science and Engineering, Hamad Bin Khalifa University, 5855, Doha, Qatar (email: yyang@hbku.edu.qa). 
		T. Huang is with Science Program, Texas A \& M University at Qatar, Doha 23874, Qatar (e-mail: tingwen.huang@qatar.tamu.edu).}

}
{}
\maketitle
\thispagestyle{fancy}
\chead{This work has been submitted to the IEEE for possible publication. Copyright may be transferred without notice, after which this version may no longer be accessible.}

\begin{abstract}
The sparse representation of graphs has shown great potential for accelerating the computation of graph applications (e.g., Social Networks, Knowledge Graphs) on traditional computing architectures (CPU, GPU, or TPU). But the exploration of large-scale sparse graph computing on processing-in-memory (PIM) platforms (typically with memristive crossbars) is still in its infancy. To implement the computation or storage of large-scale or batch graphs on memristive crossbars, a natural assumption is that a large-scale crossbar is demanded, but with low utilization. Some recent works question this assumption, to avoid the waste of storage and computational resource, the fixed-size or progressively scheduled ``block partition'' schemes are proposed. {However, these methods are coarse-grained or static, and are not effectively sparsity-aware.} This work proposes the dynamic sparsity-aware mapping scheme generating method that models the problem with a sequential decision-making model, and optimizes it by reinforcement learning (RL) algorithm (\textit{REINFORCE}). Our generating model (\textit{LSTM}, combined with the dynamic-fill scheme) generates remarkable mapping performance on a small-scale graph/matrix data ({complete mapping costs} 43\% area of the original matrix) and two large-scale matrix data ({costing} 22.5\% area on \textit{qh882} and 17.1\% area on \textit{qh1484}). Our method may be extended to sparse graph computing on other PIM architectures, not limited to the memristive device-based platforms.
\end{abstract}

\begin{IEEEkeywords}
Memristor, Sparsity, Large-scale graph, LSTM, Reinforcement learning
\end{IEEEkeywords}

\section{Introduction}
\IEEEPARstart{G}{raph} data structure is typically represented by an adjacency matrix, which has extensive sparsity \cite{song2018graphr,dai2019graphsar,song2022graph,zhao2018learning}. It relies on traditional matrix compression format, e.g., CSR, CSC, and COO, to save substantial storage resources, and is effectively processed by the sparse-based computing algorithm or library (SpMM, SpMV) on traditional computing devices (CPU, GPU, or TPU) \cite{huang2020ge}. However, in terms of {processing-in-memory} (PIM) or computation-in-memory (CIM) platforms, which are typically implemented by memristive crossbars, the graph data must be restored from the storage format to the computing format (adjacency matrix) \cite{song2018graphr} before mapping.
Directly mapping large-scale sparse graph data on a crossbar is not appropriate, for it will seriously affect the utilization and power consumption of memristive platforms.
In addition to the large-scale graph data, it is also fatal for batch graphs computing, in which the adjacency matrices are usually integrated into a large-scale super-matrix, with only the sub-graphs being internally connected, and the adjacency relationship across the graphs are null \cite{balog2019fast}. {Unfortunately, the current fabrication technology of memristor crossbars is immature, and it is difficult to fabricate large-scale memristive crossbars with {high yield}. Therefore, it is necessary to make efficient usage of the discrete small-scale crossbars on memristor-based computing platforms.}

In recent years, some PIM-based literature \cite{cui2016towards,song2018graphr,dai2019graphsar} have made effective attempts in the efficient computation of sparse graphs on neuromorphic computing platforms.
Following clustering or reordering \cite{10.1145/800195.805928} the sparse adjacency matrix, some fixed-size block partition or progressive partition schemes are proposed, among which only the blocks with non-zero entries need to be mapped to the crossbar \cite{balog2019fast, dai2019graphsar, song2018graphr}.
From our perspective, these schemes fail to propose an effective mapping scheme for the graph data, which should be feasible, scalable, dynamic, and flexible, especially in the scenario of large-scale graph or batch graphs computing. 
This paper focuses on filling this gap, which is crucial in optimizing the computing/storage efficiency and resource utilization of large-scale graphs on the memristive platforms. According to the scenarios and characteristics of memristive crossbar based computing, we first propose several principles of dynamic sparsity-aware mapping and coding framework. We then model the problem as a sequential decision-making problem, which is heuristically sampled and optimized by a policy gradient-based reinforcement learning algorithm. Eventually, our experiments show remarkable mapping results on both small-scale and large-scale graph/matrix datasets.

Overall, our contributions can be summarized as follows:
\begin{itemize}
\item \textbf{Modeling} {We simplify the mapping scheme generation problem, and further formulate it as the sequential decision-making problem.}
\item \textbf{Optimization} {We further exploit \textit{LSTM + Dynamic-fill} to model the problem, and optimize it heuristically by the reinforcement learning algorithm.}
\item \textbf{Scalability} Our method can meet the real constraint (allowable limited crossbar size, the complexity of peripheral circuits, {etc.}) of the deployed platforms, that is, it is flexible and scalable.
\end{itemize}
Our work eliminates the dependence of large-scale graph computing on integrated crossbars and allows for the utilization of discrete crossbars with limited sizes. This improves the utilization and feasibility of memristive crossbars, enabling dynamic sparsity-aware mapping in large-scale graph computing. Our code is available at \url{https://github.com/blyucs/AutoGMap}.

\section{Related work}
\textbf{Processing-in-memory and neuromorphic computing.}
Data exchange between PUs and off-chip storage devices (hard drives, flashes) consumes two orders of magnitude more energy than a floating-point operation \cite{keckler2011gpus}.
It becomes more serious for neural network applications and real-time automatic systems \cite{ghapani2017distributed,ao2018distributed,zhang2018new,zhao2021adaptive}, which substantially rely on data storage and memory exchange (both w.r.t. feature maps and weight parameters). The ``Memory Wall'' problem in von-Neumann architecture will surely become a bottleneck for progress in these areas.
Given the capacity to achieve low-power consumption and low inference latency, PIM \cite{ahn2015scalable, zhang2014top,chi2016prime} is a feasible solution to tackle these issues \cite{zhuo2019graphq,lin2019learning}, {which integrates the computation and data storage.}

\textbf{Memristor crossbar-based computation.}
The emerging studies of memristor \cite{chua1971memristor, strukov2008missing, hu2016memristive, krestinskaya2019neuromemristive} have shown its great potential on PIM platforms. With the memristors structured into the crossbar, it performs matrix-vector multiplication efficiently and has been widely studied to accelerate neural network (NN) applications.
Much more works have studied the deployment of different types of neural networks (ex-situ) w.r.t. different benchmark tasks on memristor crossbar arrays \cite{wen2018memristor, yakopcic2017extremely, yang2020retransformer}. Other works also study the training procedure of neural networks (in-situ) \cite{yan2020training, prezioso2015training, kataeva2015efficient, li2014training, cheng2017time, duan2014memristor}.
Wen et.al \cite{wen2018memristor} propose a novel memristor-based computational architecture for the Echo State Network (MESN) with the online Least Mean Square (LMS) algorithm. {Chen et.al \cite{chen2021efficient} design the memristor-based circuit to implement Fully Convolutional Networks (FCNs) for image segmentation application.}
Recently, as a newly evolutionary neural network structure, the Transformer \cite{vaswani2017attention} has also been studied under the scenario of memristor crossbar arrays \cite{yang2020retransformer}. {Our work is also partially motivated by \cite{chen2021efficient} and \cite{veluri2021low}, which are concerned with the acceleration of the hardware implementation of artificial neural networks.}

\textbf{Resource-aware computation and energy saving.}
Recently, much more attention has been paid to the resources and energy consumption of computing. Some notable works on resource-aware computation optimization and software/hardware co-design models \cite{lyu2021multiobjective, lyu2021resource} are proposed.
Targeting the resource consumption and energy saving on neuromorphic computing platforms, some works \cite{wen2019memristor, wen2020ckfo} propose some effective ways for accelerating the memristor-based CNNs on classification or segmentation tasks.

\textbf{Graph data processing and acceleration.}
Towards the sparse matrix-vector multiplication based on memristor crossbar, Cui et.al. \cite{cui2016towards} propose an improved/generalized reordering algorithm based on Cuthill-McKee reordering to reduce the bandwidth of graph adjacency matrix, thus improving the efficiency of the crossbars{, whereas no} blocks mapping scheme is studied after reordering.
Balong et.al. \cite{balog2019fast} directly utilize the Cuthill-McKee reordering algorithm followed by three densified diagonal-blocks coverage, to speed up the training efficiency, thus demonstrating the competitive efficiency on TPUv2 (dense acceleration hardware) to GPU (sparse acceleration hardware). 
Targeting the heterogeneous accelerators for graph computation on PIM architecture, some works \cite{song2018graphr, dai2019graphsar} propose sparse graph partition schemes to improve the storage efficiency.
GraphR \cite{song2018graphr} statically partitions adjacency matrix, and {combines} sparse compression format, to reduce memory consumption. On the WikiNote dataset, it cost only 0.2\% of the original size when combined with the COO representation. GraphSAR \cite{dai2019graphsar} proposes the sparsity-aware partition scheme to store the large-scale sparse matrix. After {partitioning} the adjacency matrix into fixed-small-size (e.g. $8 \times 8$), it only directly stores the blocks with a non-zero density larger than 0.5, otherwise further divides the blocks into $4 \times 4$. By this means, the large-scale graphs data are stored in the form of small matrices, thus greatly increasing the utilization of the discrete fragmentary crossbars.
Although these works endeavor to improve the computation efficiency and resource utilization by block partition scheme, no dynamic and real sparsity-aware partition methods are proposed.

\section{Preliminary}
In parallel computing on conventional hardware (CPU, GPU), the Intel MKL, NVIDIA cuBLAS library, cuSparse, and other libraries directly implement SpMV based on the compressed format.
However, for large-scale graph data or batch graphs, although the memristor-based crossbar operation utilizes a process-in-memory architecture to reduce the complexity of matrix-vector multiplication from $O(n^2)$ to $O(1)$, it is clear that the complete mapping of the integrated matrix to the crossbar is unwise. 
Taking spectral-based GCN \cite{kipf2016semi} as an example, the layer-wise propagation as Eq. \eqref{eq:Z_l}:
\begin{equation}
\begin{array}{ll} \label{eq:Z_l}
Z_{l+1}=\sigma(\hat{D}^{-1/2}\hat{A}\hat{D}^{-1/2}Z_{l}W_{l})
\end{array}
\end{equation}
It involves the whole sparse adjacency matrix, so complete mapping of graphs on crossbar/hypercube is very resource-consuming.
A preferable way is to divide the sparse matrix into blocks, and only the blocks containing non-zero elements need to be mapped \cite{cui2016towards, balog2019fast}. However, the non-zero elements of the sparse matrix with graph structure are generally scattered, which will increase the complexity of peripheral circuits and communication between sub-crossbars (blocks) \cite{cui2016towards}. Targeting the goal of ``communication optimal'' \cite{cui2016towards}, the communication of the blocks in the same row needs to be minimized, thus reducing the complexity of the peripheral circuit. 
To achieve this, non-zeros need to be located closer enough and preferably distributed around with the diagonal.

\begin{figure}[!ht]
\centering
\includegraphics[width=8.5cm]{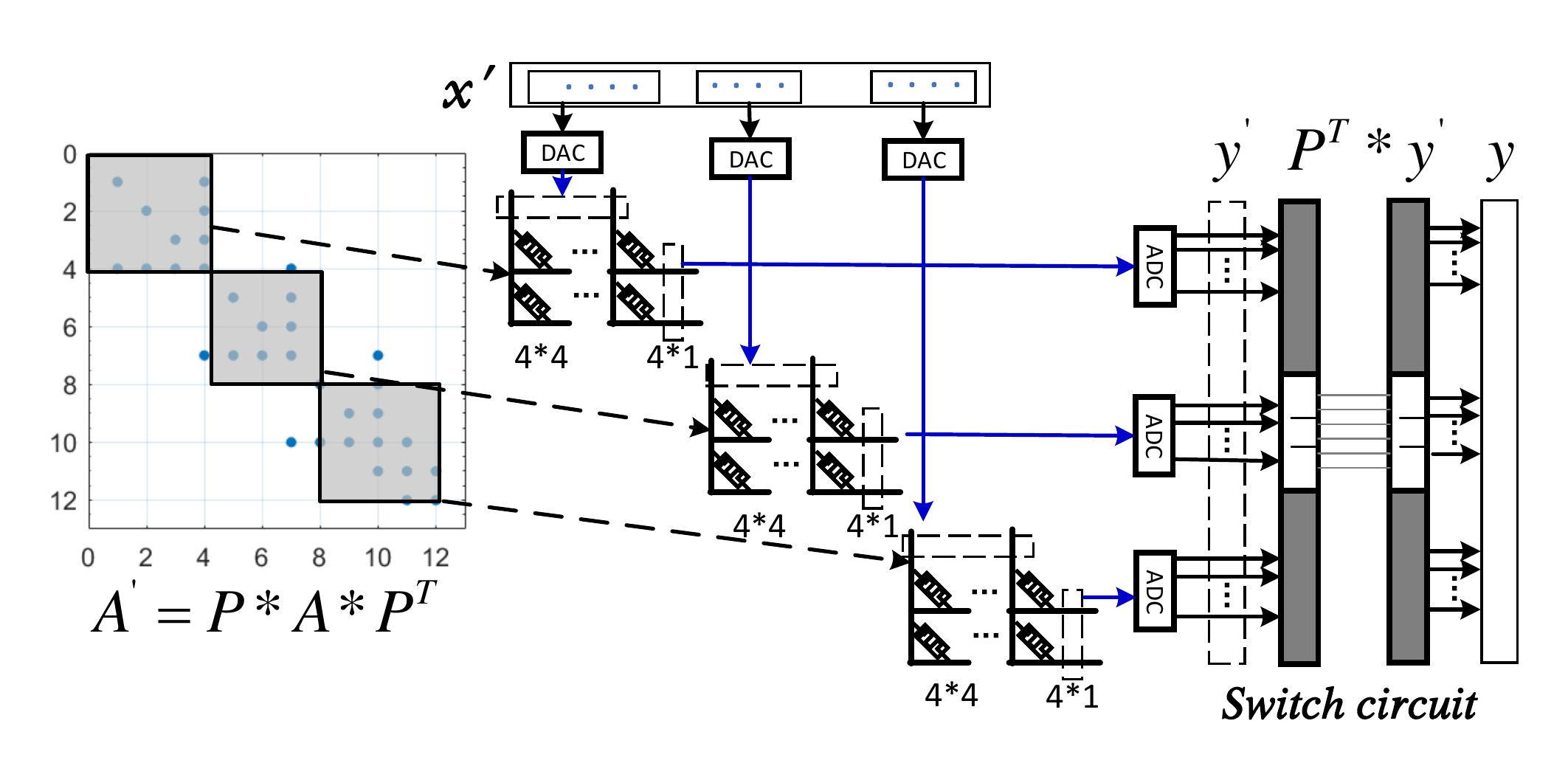}
\caption{Diagram of matrix-vector multiplication propagation ($y=Ax$) with the matrix reordering method. The matrix $A^{'}$ is programmed into a batch of small-scale crossbars, and the transformed vector $x^{'}$ serves as the inputs of the crossbars. Finally, the switch circuit is resorted to realize the reverse transformation $y=P^{T}y^{'}$.}
\label{fig:diag_mapping}
\end{figure}

Matrix-vector multiplication is the common atomic operation of current Artificial Intelligence applications (e.g. CNNs \cite{lecun1998gradient, simonyan2014very, he2016deep}, RNNs \cite{hochreiter1997long}, GNNs \cite{hamilton2017inductive, kipf2016semi, velivckovic2017graph}), and {consists} of the multiplication and accumulation.
In the computation of the memristive crossbar, multiplication is implemented by Ohm's law, and accumulation is implemented by Kirchhoff’s Current Law in analog domain \cite{Dot-Product-Engine}.
Fig. \ref{fig:diag_mapping} showcases the circuit simulation process of a matrix-vector multiplication propagation ($y = Ax$) by reordering and its corresponding post-transformation.
Supposing the original calculation of the matrix-vector multiplication is:
\begin{equation}
y=Ax
\end{equation}
By Cuthill-Mckee reordering, the matrix $A$ is transformed to:
\begin{equation}
A^{'}=PAP^{T}
\end{equation}
which is feasible to be deployed on the crossbar with fewer cost \cite{cui2016towards}. To ensure the calculation rules of block matrix multiplication, the input vector needs the transformation, denoted as:
\begin{equation}
x^{'}=Px
\end{equation}
The calculation after the transformation is represented as:
\begin{equation}
y^{'}=A^{'}x^{'}=PAP^{T}Px=PAx=Py
\end{equation}
Then, after the calculation of crossbars, the compositive output vector $y^{'}$ needs to be reversely transformed back to $y$, denoted as:
\begin{equation}
y=P^{T}y^{'}
\end{equation}
which may be realized by the switch circuit during the hardware design.

\begin{figure}[!ht]
\centering
\includegraphics[width=8.5cm]{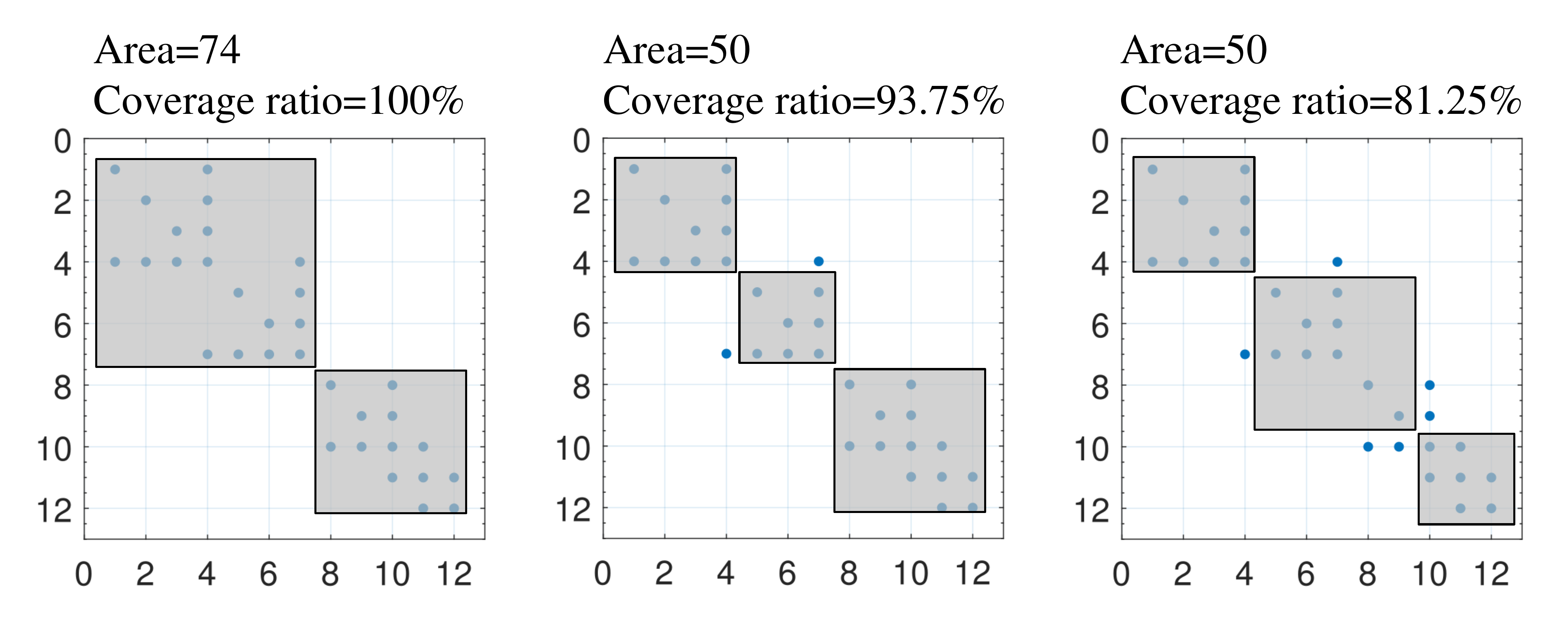}
\caption{The comparison of coverage ratio and blocks area (cost) under different mapping schemes. {To realize the complete mapping (coverage), the mapping blocks generally cost much more area, as the left scheme shows. But the middle and rights schemes are infeasible for the deployment, which fails to reach the complete coverage.}}
\label{fig:preli}
\end{figure}

{More acceleration and deployment background is proposed in \cite{lyu2022efficient}.}
Considering only diagonally connected block partition, as shown in Fig. \ref{fig:preli}, different block schedule schemes come up with different coverage ratio and total area (cost). Therefore, the question worth exploring is how to generate the best mapping scheme (complete coverage) with minimum cost.

\section{Problem Formulation}
For simplicity, we first consider the formulation of the mapping scheme with only diagonal blocks. The optimization objective is to maximize the expectation of the reward function of the candidate scheme, as Eq. \eqref{eq:s_block}:
\begin{equation}
\begin{array}{ll} \label{eq:s_block}
\mathop{argmax}\limits_{\mathop{s}}& f(s_{1}, s_{2}, s_{3}, ..., s_{n})\\
\textit { s.t. }& \sum\limits_{n=1}^{n}s_{n} = N\\
& n \leq N
\end{array}
\end{equation}
{where $f$ is the reward function of one candidate mapping scheme, and $s$ is the ``size'' vector of the sequential blocks, $N$ is the diagonal size.}
In terms of the solution space, this problem can be analogous to a typical pure numerical problem, the integer factorization by addition, that is, given a positive integer $n$, the number of the factorization schemes is $2^{N-1}$. So the algorithm complexity of the violent solution is $O(2^N)$. Considering a large numerical value $N$ (e.g. $100$, $1000$), solving this problem violently is obviously computationally impossible.
Additionally, it is infeasible to formulate this problem as the integer programming or the discrete optimization problem, for the variable number is uncertain.

We endeavor to transform the expression of the solution space, instead of using the size variables with an uncertain number, we resort to {$N-1$ variables ($0/1$)} to form the solution space. This may be detailed interpreted as that there exist $N-1$ decision points on the diagonal of the adjacency matrix, and the action space of each point is {0: Start a new block, 1: Continued to expand the previous block}, represented by $x_{i}$.
Then the optimization problem is transformed to Eq. \eqref{eq:zero-one}:
\begin{equation}
\begin{array}{ll} \label{eq:zero-one}
\mathop{argmax}\limits_{\mathop{x}}& f(p(x_{1}, x_{2}, x_{3}, ..., x_{N}));\\
\textit { s.t. }& \ x_{i} \in {\{0, 1\}}
\end{array}
\end{equation}
where $p$ is the parse function that transfers 0-1 decision variable $x$ to the block size vector $s$.
Although the size of the solution space is the same with Eq. \eqref{eq:s_block}, $2^{N-1}$, consistent with the 0-1 nonlinear integer programming problem, it is easier to formulate this problem as a programming problem. Unfortunately, it still can not be solved by optimization methods, for the definition of the optimization objective is not a continuous high-dimensional hyperplane, but regarding the distribution of nonzero elements of the matrix, which is in a discrete space.
\begin{figure}[!ht]
\centering
\includegraphics[width=8.5cm]{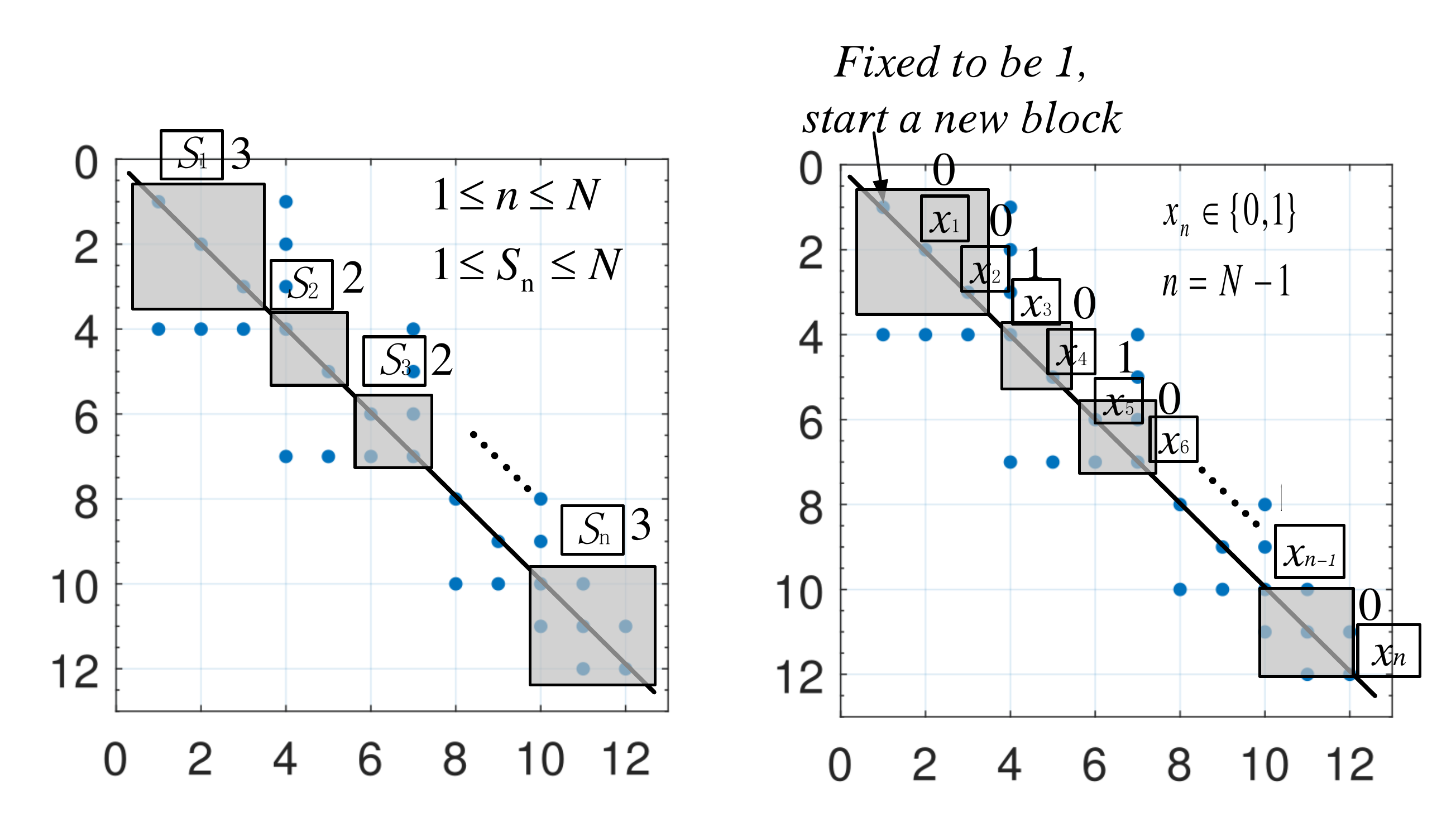}
\caption{Left: Illustration of the optimization problem defined in Eq. (2). The block size is the optimization variable, but the number of which is not determined. Right: Illustration of the optimization problem defined in Eq. (3). Each grid point on the diagonal needs a decision indicating whether to start a new block or continue the frontier one, the variable number is determined.}
\label{fig:decision}
\end{figure}
Based on the background of the crossbar and the formulated problem, we put forward the basic principle of mapping large-scale sparse graphs on memristive crossbars:
\begin{itemize}
\item Complete coverage capability, valid for arbitrary non-zeros distribution, and do not exceed the whole area.
\item No overlaps between blocks.
\item Adaptable to the deployment and compile system, that means, coding rules should be simple for the circuit design.
\item {Least cost of crossbar area.}
\end{itemize}

\section{Methodology}
\subsection{Modeling}
\textbf{Modeling diagonal-blocks.} 
{Based} on the problem formulation and the background of crossbar-based computation (Fig. \ref{fig:diag_mapping}), optimization variables defined in Eq. \eqref{eq:zero-one} are pre-and-post related, we may treat it as a sequential {generation} problem and further leverage the classical LSTM networks {as the sampling controller.}
For each time-step ($t$, stands for the decision points number), the output embedding data models the representation of each action and serves as the input vector of the next time-step, {and the propagation of LSTM cell calculates} as the Eq. \eqref{eq:lstm_begin}-\eqref{eq:lstm_end}:
\begin{equation} \label{eq:lstm_begin}
\begin{aligned}
f_{t}=\sigma(W_{f}[h_{t-1},x_{t}]+b_{f})
\end{aligned}
\end{equation}
\begin{equation} \label{eq:3}
\begin{aligned}
i_{t}=\sigma(W_{i}[h_{t-1},x_{t}]+b_{i})\\
\end{aligned}
\end{equation}
\begin{equation} \label{eq:4}
\begin{aligned}
g_{t}=tanh(W_{g}[h_{t-1},x_{t}]+b_{g})\\
\end{aligned}
\end{equation}
\begin{equation} \label{eq:5}
\begin{aligned}
o_{t}=\sigma(W_{o}[h_{t-1},x_{t}]+b_{o})\\
\end{aligned}
\end{equation}
\begin{equation} \label{eq:6}
\begin{aligned}
c_{t}=f_{t}*c_{t-1}+i_{t}*g_{t}\\
\end{aligned}
\end{equation}
\begin{equation} \label{eq:lstm_end}
\begin{aligned}
h_{t}=o_{t}*tanh(c_{t})\\
\end{aligned}
\end{equation}
where $W$ denotes the weights and $b$ represents the bias.
At time-step $t$, $f_t$, $i_t$, $g_t$, $o_t$ represents the forget gate, input gate, cell state, and output gate, respectively. $h_{t-1}$ and $x_t$ signify the hidden state at time $t-1$ and the input at time $t$, respectively.
The sequential generation technique can be formulated as a classification problem and accomplished by a fully connected network that takes the LSTM cell's output embedding vector (per time-step) as input.
\begin{equation} \label{eq:8}
\begin{aligned}
p^t=softmax(W^t_{fc}*out_{t}+b^{t})\\
\end{aligned}
\end{equation}
The output indices that represent the index/operations are sampled according to the multinomial distribution $(p^t_{1}, \cdot\cdot\cdot, p^t_{C})$, where $C$ is {the classification number of FC. The sampling details are} presented in Algo. \ref{algo:sampling}.

\begin{figure}[!ht]
\centering
\includegraphics[width=7.5cm]{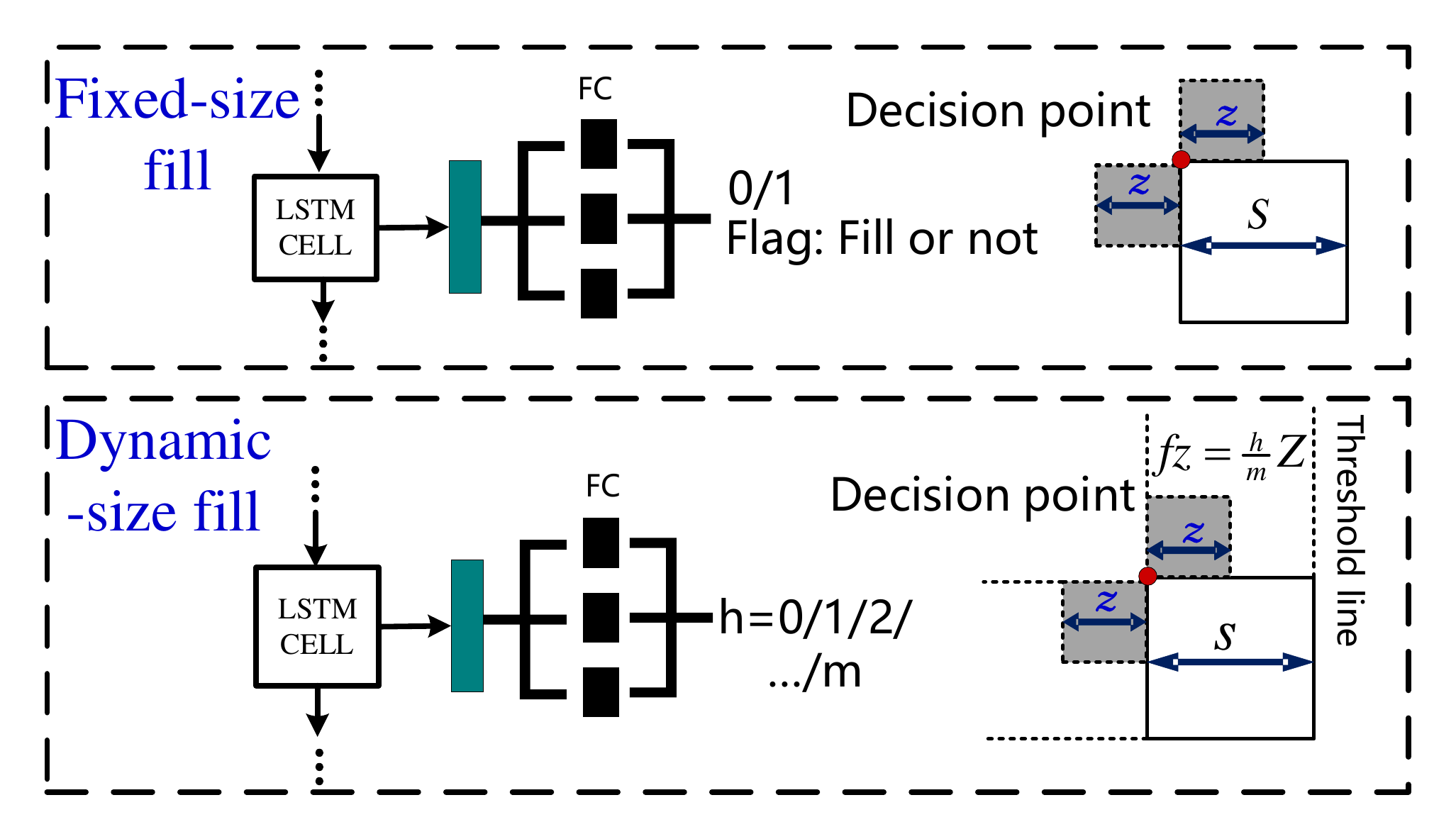}
\caption{Up: Fill the gaps with fixed-size blocks, with the fully-connected network model the binary classification problem, the output value means to fill or not, as Eq. (11). Down: Fill the gaps with dynamic size blocks, with the fully-connected network model the multi-classification problem, as Eq. (12), the classification output stands for the portion of current fill-block size, e.g., indices $[0, 1, 2, 3, 4, 5]$ stands for the ratio $[0, 1/5, 2/5, 3/5, 4/5, 1]$.}
\label{fig:dynamic_size}
\end{figure}

\textbf{Modeling fill-blocks.}
As shown in Fig. \ref{fig:preli}, schemes with only diagonal-blocks are not enough to provide the feasible coverage solution, for the joints of two adjacent blocks are blind areas. We heuristically employ the methods in work \cite{balog2019fast}, in which ``fill the gaps'' blocks are distributed on two sides of the diagonal. The difference is that our solution is dynamic rather than static.
Intuitively, we may model the decision of the ``fill the gaps'' blocks the same as the diagonal-blocks. Thus the optimization formula is presented in Eq. \eqref{eq:zero-one-fill}, in which the FCs serve as the binary classifiers to determine whether to fill the gap with fixed-size or not. Unfortunately, this strategy will inevitably come up with a waste of resources (memristors, energy). To address this, we creatively propose a dynamic-fill scheme to prevent this limitation and further improve the utilization. In this strategy, the FCs serves as the multi-classifier, and the output classification value stands for the fill-block size, in the form of a proportion of the current diagonal-block. The illustration of two schemes (fixed-size fill and dynamic size fill) is shown in Fig. \ref{fig:dynamic_size}. {Since only when the diagonal-block decision is ``to start a new block'', the corresponding fill-block needs to be decided, the fill-block sequence needs to be masked by the diagonal-block sequence.}
\begin{equation}
\begin{array}{ll} \label{eq:zero-one-fill}
\mathop{ argmax}\limits_{\mathop{x, z}}& f(p(x, z));\\
\textit { s.t. }&  x_{i} \in {\{0, 1\}}\\
&  z_{i} \in {\{0, 1\}}
\end{array}
\end{equation}

\begin{equation}
\begin{array}{ll} \label{eq:dynamic-fill}
\mathop{ argmax}\limits_{\mathop{x, z}}& f(p(x, z));\\
\textit { s.t. }&  x_{i} \in {\{0, 1\}}\\
&  z_{i} \in {\{0, 1/m, 2/m, ..., 1\}}
\end{array}
\end{equation}

\begin{figure}[!ht]
\centering
\includegraphics[width=7.5cm]{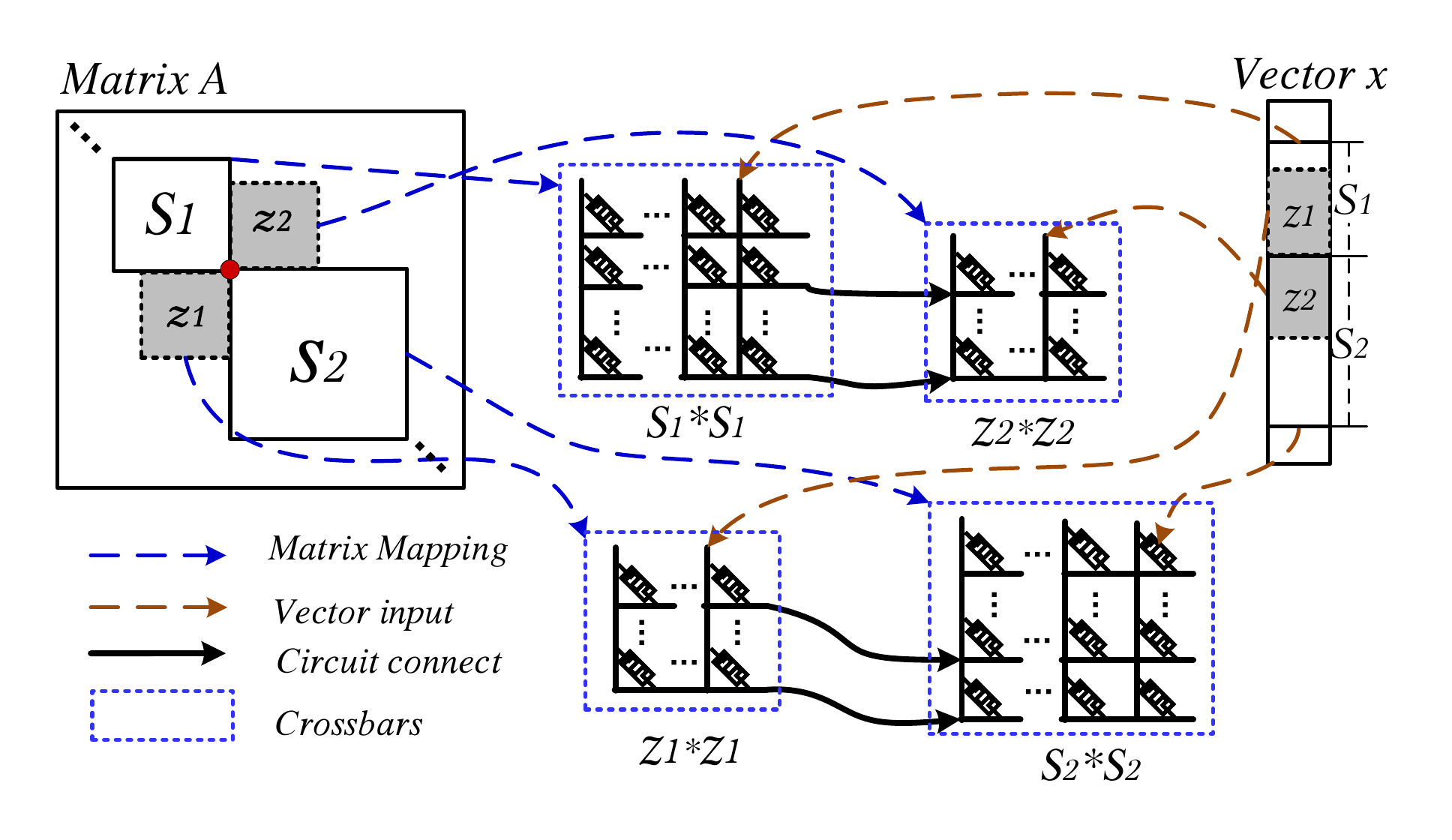}
\caption{{Illustration of the adjacency matrix mapping onto the crossbar for matrix-vector multiplication. The diagonal and fill-blocks are mapped onto the allowable small-scale crossbars. According to the Kirchhoff’s Current Law, blocks in the same row are connected, and the corresponding sub-vector serves as the input of the crossbar, respectively.}}
\label{fig:diag_fill_map}
\end{figure}

Taking matrix-vector multiplication ($y=Ax$) as an example, Fig. \ref{fig:diag_fill_map} showcases the mapping of graph data on crossbar. Diagonal and fill-blocks are mapped to the allowable small-scale crossbars. {Based} on the Kirchhoff’s Current Law, blocks in the same row are connected, and the corresponding sub-vector (splitting follows the rule of ``block matrix multiplication'') is the data input of the crossbar, respectively. In this way, the allowable small-scale crossbars can be effectively utilized. 

\subsection{Optimization}
{The optimization objective is to maximize the expected reward of the sampled schedule scheme, and the reward function is not differentiable with respect to the controller parameters, so the policy gradient algorithm is used to construct the gradient estimates of the controller's weight parameters.}
The issues that need to be addressed are as follows:
\begin{itemize}
\item In such a huge action space, how to sample the block scheduling scheme directionally and heuristically.
\item Reward the sampled schedule scheme efficiently and timely, which should be instructive to the training.
\item When evaluating, multiple metrics need to be considered, including the coverage ratio, block area, which could be contradictory, as well as the trade-off between the two. 
\end{itemize}
\begin{table}[!htb]
\centering
\caption{The RL Notation and definition of current problem.}
\setlength{\tabcolsep}{0.5mm}{
\begin{tabular} {p{70pt}p{150pt}}
\toprule
    Notation & Definition\\
\toprule
Environment ($e$) & Original matrix $A$\\
Reward ($R$) & $f(p(x,z))$\\
Agent ($\theta$) &  LSTM and FCs\\
Action ($a$) & {Sequentially sampled decision indices}\\
Action space & Fixed-fill: $2^{N}$, dynamic-fill: $2^{N-1}*(m+1)^{N-1}$\\
\toprule
\end{tabular}
}
\centering
\label{table:notation}
\end{table}
Further, the reinforcement procedure {is formulated as} to optimize the $\theta$ to achieve the optimal reward ($R$) of the candidate mapping scheme, as Eq.\eqref{eq:theta}:
\begin{equation} \label{eq:theta}
\begin{aligned}
\theta^{*} &= \mathop{\arg\max}_{\theta}\mathcal{J}(\theta) = \mathop{\arg\max}_{\theta}\mathbb{E}_{\pi(a_{1:T};\theta)}(R)
\end{aligned}
\end{equation}
where $\mathcal{J}(\theta)$ is the optimization objective function, represented by the expectation of the reward achieved by coverage sequence action $a_{1:T}$, where $T$ is the length of action vector.
\begin{equation} \label{eq:theta1}
\begin{aligned}
\nabla_{\theta}{\mathcal{J}(\theta)}&=\sum_{t=1}^{T}\mathbb{E}_{\pi(a_{1:T};\theta)}[R\nabla_{\theta}\log(\pi(a_{t}|a_{1:t-1};\theta))]\\
\end{aligned}
\end{equation}
the gradient value is approximately calculated by sampling, that is:
\begin{equation}
\begin{aligned}
\nabla_{\theta}{\mathcal{J}(\theta)} \approx \frac{1}{M}\sum_{m=1}^{M} \sum_{t=1}^{T}\mathbb{E}_{\pi(a_{1:T};\theta)}[R_{m}\nabla_{\theta}\log(\pi(a_{t}|a_{1:t-1};\theta))]\\
\end{aligned}
\end{equation}
the algorithm of the training of the agent is presented in Algo. \ref{algo:tainagent}.
As for the reward $R_m$, it involves two objectives: 
\begin{itemize}
\item Coverage ratio, which relates to the performance of the computation on the crossbar.
\item Area cost, which relates to memristor and energy consumption.
\end{itemize}
Distinctly, there exists the contradiction between the two, that is we attempt to maximize the coverage ratio (or even complete coverage) and minimize the area cost of crossbars. To achieve this, we leverage the single-policy MORL \cite{1998multi-criteria,mannor2004a,moffaert2013scalarized} which resort to the scalarization function to transform the multi-objective problem into a standard single-objective one. The scalarization function $f$ projects an objective vector $\mathbf{v}$ to a scalar one:
$v_{\mathbf{w}}=f(\mathbf{v}, \mathbf{w})$
where $\boldsymbol{w}$ is a weight vector parameterizing $f$. We employ the simple way of weight-sum as the scalarization function, as Eq. \eqref{eq:multi-objective}:
\begin{equation} \label{eq:multi-objective}
\begin{aligned}
R_{m} = a * Coverage(x, z) + (1-a) * Area(x, z)
\end{aligned}
\end{equation}
{where $a$ is the harmonic coefficient. In general, the application scenario requires complete coverage (coverage ratio is 1), consequently, what's serious is the crossbar's consumption (area). In such a large action space, an extremely large proportion of the candidate schemes is capable to reach complete coverage, but with different area ratios. So if we set $a$ to be 1 or close to 1 from the beginning, the agent would tend to sample the complete coverage schemes with high probability, which even with different costs, the reward keeps consistent, thus it makes the gradient of the controller disappear. Consequently, we must have the weighted-sum coefficient $a$ serves as the hyperparameter that is empirically tuned.}
In addition, the balance of exploration and exploitation is achieved by probabilistic sampling. 
\begin{algorithm}
\label{algo:sampling}
\caption{Agent sampling.}
\LinesNumbered
\KwIn{Null}
\KwOut{$diagonal\_actions,fill\_actions,log\_prob$}
$inputs, hidden$ = random\_initialize\;
\For{ $i \gets 0 \to action\_len-1$}{
$output, hidden$ $\gets$ LSTM($inputs, hidden$)\;
{logits $\gets$ the $i$th diagonal fcs output\;
$softmax\_logits$ $\gets$ softmax($logits$)\;
$d\_action$ $\gets$ multinomial sampling by softmax\_logits };
$cur\_log\_prob$ $\gets$ -nll\_loss(log($softmax\_logits$), $d\_action$)\;
$log\_prob = log\_prob + curr\_log\_prob$\;
$diagonal\_actions$ append $d\_action$\;
$inputs \gets output$\;
\
\CommentSty{/*"Fill" masked by "Diagonal", 0: Start a new block*/}\;
\If {$d\_action == 0$}{
$output, hidden$ $\gets$ LSTM($inputs, hidden$)\;
{logits $\gets$ the $i$th fill fcs output\;
$softmax\_logits$ $\gets$ softmax($logits$)\;
$d\_action$ $\gets$ multinomial sampling by softmax\_logits }; 
$cur\_log\_prob$ $\gets$ -nll\_loss(log($softmax\_logits$), $f\_action$)\;
$fill\_actions$ append $f\_action$\;
$log\_prob \gets log\_prob + curr\_log\_prob$\;
$inputs \gets output$\;
}
}
\end{algorithm}

\begin{algorithm}
\label{algo:tainagent}
\caption{Optimizing the sampling agent (REINFORCE with baseline).}
\LinesNumbered
\KwIn{$log\_prob$, $Reward$}
\KwOut{Null}
$baseline \gets decay*baseline + (1-decay)*reward$\;
$adv \gets reward - baseline$\;
loss $\gets$ $-log\_prob * adv$\;
loss.backward() \CommentSty{/*Gradients ($loss$ is differentiable w.r.t. the parameter weights of the $LSTM$ and the $FCs$) calculated by $autograd$*/}\;
Gradients applied \CommentSty{/*Gradient descend by the optimizer.*/}\;
\end{algorithm}

\begin{algorithm}
\label{algo:AutoGMap}
\caption{AutoGMap.}
\LinesNumbered
\KwIn{$A(Matrix), agent\_config, a$}
\KwOut{$digonal\_blks, fill\_blks$}
Agent $\gets$ create\_agent(agent\_config)
\For{$epoch \gets 0 \to num\_epoch-1$}{
$diagonal\_action,fill\_action,log\_prob$ $\gets$ Agent.sample()\;
$diagonal\_blks$ $\gets$ parse\_d($diagonal\_action$)\;
$fill\_blks$ $\gets$ parse\_f($fill\_action$)\;
$C\_ratio$ $\gets$ C\_cal($diagonal\_action,fill\_action$)\;
$A\_ratio$ $\gets$ A\_cal($diagonal\_action,fill\_action$)\;
$Reward \gets a*C\_ratio + (1-a)*A\_ratio$\;
train\_agent($log\_prob$, $Reward$)\;
}
\end{algorithm}

\subsection{Overall algorithm}
{
Our whole structure of the method is presented in Algo.\ref{algo:AutoGMap}.
The overall diagram of our modelling and optimization method is shown in Fig. \ref{fig:LSTM_RL}. 
\begin{figure}[!ht]
\centering
\includegraphics[width=8.5cm]{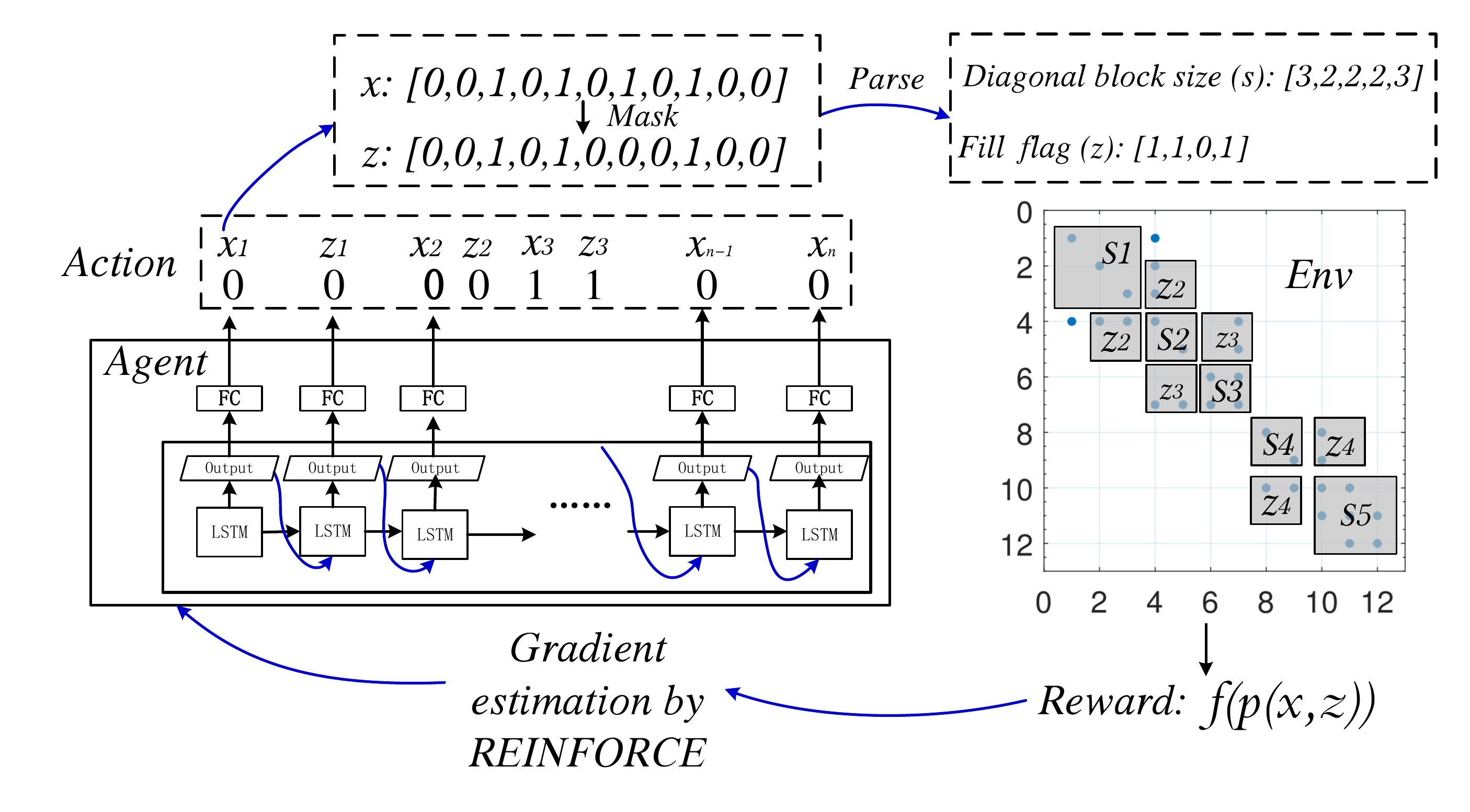}
\caption{{LSTM sequentially samples the action vectors $x, z$, which represent the mapping scheme, followed by the mapping scheme parsing and the evaluation according to non-zero elements distribution in the original matrix. After evaluation, the reward is exploited to constitute the gradient estimation value of the agent (LSTM + FCs).}}
\label{fig:LSTM_RL}
\end{figure}
}

\subsection{Relation and comparisons with other works}
In \cite{cui2016towards}, a generalized reordering method to reduce the bandwidth of the matrix \cite{10.1145/800195.805928} is proposed, which can handle any asymmetrical rectangle matrix, compared with the original Cuthill-McKee reordering algorithm that can only handle the symmetrical squares. However, this work does not address the mapping scheme and its optimization. Our work does not focus on the innovation and generalization of the reordering algorithm, but rather on the automatically generating an efficient and reasonable deployment scheme after reordering a large-scale sparse matrix.
In GraphR \cite{song2018graphr}, the adjacency matrix of a graph is partitioned into four sub-graphs with the consideration of the sparsity, the method in GraphSAR \cite{dai2019graphsar} remains the same pattern that even is progressively partitioned. Their partition scheme is fixed, whereas our work makes the scheme flexible and scalable by intelligent generating.
In work \cite{balog2019fast}, to improve the training efficiency of sparse graph neural networks on TPU device (sparse hardware), a batch of diagonal-blocks and two additional batches of blocks to ``fill the gap'', in which the size of the block keep consistent and the coverage schemes are fixed without considering of the distribution of the non-zero elements. Our overall mapping framework is similar to \cite{balog2019fast} (a batch of blocks arranged along with the diagonal, and two batches fill the gaps), what's the difference is that our method is dynamic and sparsity-aware, with the consideration of the adaptability to the compilation and deployment system.
Overall, compared with these related works, we adopt an intelligent generating method to achieve dynamic sparsity-aware mapping, which compares favourably with these previous works.

\section{Experiment}
\textbf{Environment.} We conduct our experiments using PyTorch $1.0$ framework on Intel CPU. In terms of the policy gradient training in reinforcement learning, our realization relies on PyTorch's Autograd mechanism to backwardly update the parameter weights (policy).

\textbf{Dataset.}
We first resort to small-scale graph data, an adjacency matrix ($22 \times 22$) that numbered 5828 in a chemical molecular dataset \textit{QM7} \cite{blum2009970,rupp2012fast}. In terms of the large-scale dataset, we experiment on two large-scale symmetric matrices, \textit{qh882} ($882 \times 882$) and \textit{qh1484} ($1484 \times 1484$). In our experiments, the matrices are reordered to lower-bandwidth symmetric matrices by Cuthill-McKee reordering algorithm as the pre-processing.

\begin{figure}[!ht]
\centering
\includegraphics[width=8.5cm]{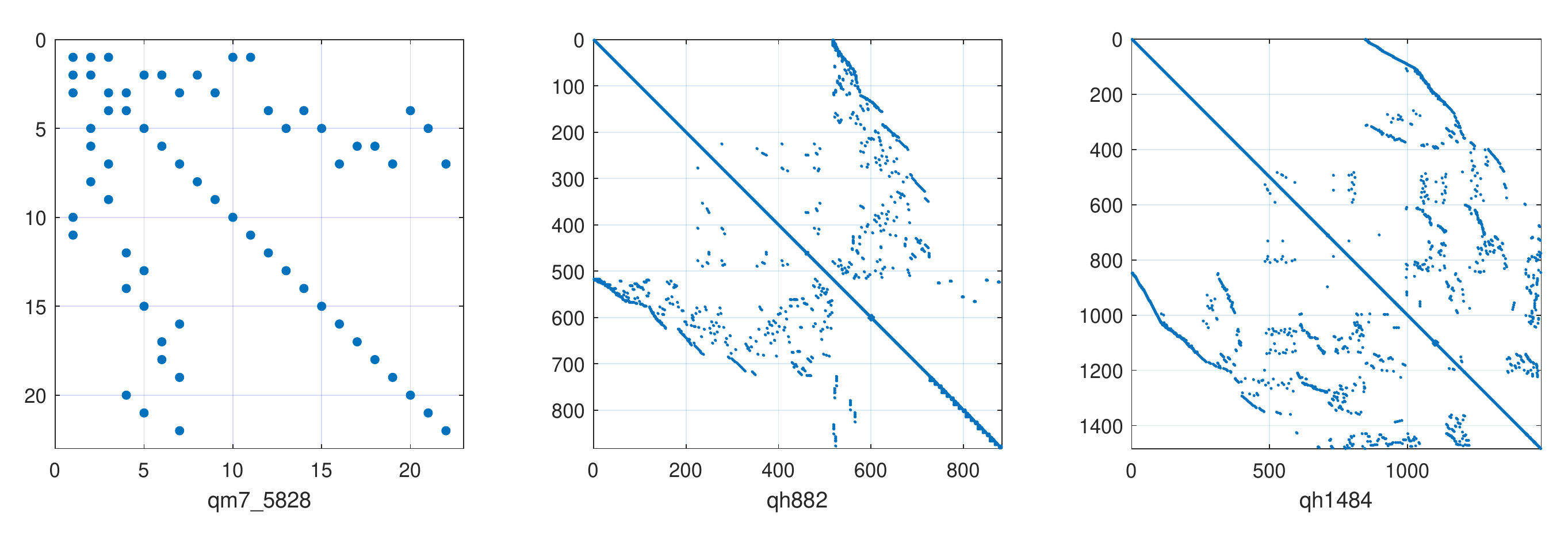}
\caption{Visualization of the datasets: \textit{QM7-5828} ($22\times22$), \textit{qh882} ($882\times882$), and \textit{qh1484} ($1484\times1484$).}
\label{fig:dataset_visual}
\end{figure}

\textbf{Metrics.}
$C\_ratio$, represents the coverage ratio of the non-zero elements of the scheme, defined as:
\begin{equation}
\begin{aligned}
C\_ratio = \frac{Nonzero\_count_{mapped\ blocks}}{Nonzero\_count_{original}}
\end{aligned}
\label{eq:c_ratio}
\end{equation}
it is scaled to be $[0, 1]$, and we attempt to reach 1 in our experiment, which means all non-zero elements must be mapped.
$A\_ratio$, the area ratio of the mapping blocks to the original matrix, defined as:
\begin{equation}
\begin{aligned}
A\_ratio = \frac{Area_{mapped\ blocks\ }}{Area_{original}}
\end{aligned}
\label{eq:a_ratio}
\end{equation}
it also stays in $[0, 1]$, the lower the better of the candidate schemes.
$Sparsity$, means the utilization of the mapped crossbar of the scheme, may be viewed as the comparison metric with the original sparsity, defined as:
\begin{equation}
\begin{aligned}
Sparsity = \frac{Nonzero\_count_{mapped\ blocks}}{Area_{mapped\ blocks}}
\end{aligned}
\label{eq:sparsity}
\end{equation}

\begin{table*}[]
\caption{Comparison and ablation study results on small-scale graph \textit{QM7-5828}.}
\label{table:R5828}
\begin{tabular}{l|l|l|l|l|l|l|l|l|l|l}
\toprule
\multicolumn{1}{c|}{\multirow{2}{*}{Methods}} & \multicolumn{1}{c|}{\multirow{2}{*}{\begin{tabular}[c]{@{}c@{}}Block\\ size\end{tabular}}} & \multirow{2}{*}{\begin{tabular}[c]{@{}l@{}}Grid\\ size\end{tabular}} & \multirow{2}{*}{\begin{tabular}[c]{@{}l@{}}Fill size/\\ grades num\end{tabular}} & \multicolumn{2}{l|}{Reward ratio} & \multicolumn{2}{c|}{Schemes} & \multicolumn{1}{c|}{\multirow{2}{*}{\begin{tabular}[c]{@{}c@{}}Coverage\\ ratio\end{tabular}}} & \multicolumn{1}{c|}{\multirow{2}{*}{\begin{tabular}[c]{@{}c@{}}Area\\ ratio\end{tabular}}} & \multirow{2}{*}{\begin{tabular}[c]{@{}l@{}}Sparsity$^\dagger$\end{tabular}} \\ \cline{5-8}
\multicolumn{1}{c|}{} & \multicolumn{1}{c|}{} &  &  & \multicolumn{1}{c|}{a} & \multicolumn{1}{c|}{1-a} & \multicolumn{1}{c|}{Diagonal-blocks size} & \multicolumn{1}{c|}{Fill-blocks size} & \multicolumn{1}{c|}{} & \multicolumn{1}{c|}{} &  \\ \hline
\multirow{3}{*}{Vanilla} & 4 & \multirow{3}{*}{/} & \multirow{3}{*}{/} & \multirow{3}{*}{/} & \multirow{3}{*}{/} & {[}4, 4, 4, 4, 4,2{]} & \multirow{3}{*}{/} & 0.5 & 0.174 & 0.620 \\ \cline{2-2} \cline{7-7} \cline{9-11} 
 & 6 &  &  &  &  & {[}6, 6, 6, 4{]} &  & 0.531 & 0.256 & 0.726 \\ \cline{2-2} \cline{7-7} \cline{9-11} 
 & 8 &  &  &  &  & {[}8, 8, 6{]} &  & 0.813 & 0.339 & 0.683 \\ \hline
\multirow{2}{*}{Vanilla+Fill} & 4 & \multirow{2}{*}{/} & 4 & \multirow{2}{*}{/} & \multirow{2}{*}{/} & {[}4, 4, 4, 4,4,2{]} & {[}1, 1, 1, 1, 1, 1{]} & 0.938 & 0.445 & 0.721 \\ \cline{2-2} \cline{4-4} \cline{7-11} 
 & 6 &  & 6 &  &  & {[}6, 6, 6, 4{]} & {[}1, 1, 1, 1{]} & 1.0 & 0.62 & 0.787 \\ \hline
\multirow{2}{*}{LSTM+RL} & \multirow{2}{*}{/} & \multirow{2}{*}{2} & \multirow{2}{*}{/} & 0.6 & 0.4 & {[}8, 2, 12{]} & \multirow{2}{*}{/} & 0.875 & 0.438 & 0.735 \\ \cline{5-7} \cline{9-11} 
 &  &  &  & 0.8 & 0.2 & {[}8, 14{]} &  & 0.938 & 0.537 & 0.769 \\ \hline
\multirow{6}{*}{LSTM+RL+Fill} & \multirow{6}{*}{/} & \multirow{6}{*}{2} & 2 & 0.8 & 0.2 & {[}8, 12, 2{]} & {[}0, 1{]} & 0.938 & 0.455 & 0.727 \\ \cline{4-11} 
 &  &  & 4 & 0.8 & 0.2 & {[}2, 2, 4, 2, 6, 4, 2{]} & {[}0, 1, 0, 1, 1, 1{]} & 0.969 & 0.388 & 0.670 \\ \cline{4-11} 
 &  &  & 4 & 0.9 & 0.1 & {[}4, 12, 4, 2{]} & {[}1, 1, 1{]} & 1.0 & 0.521 & 0.746 \\ \cline{4-11} 
 &  &  & {6} & {0.9} & {0.1} & {{[}8, 2, 10, 2 {]}} & {{[}1, 1, 1{]}} & {1.0 } & {0.537 } & {0.754 } \\ \cline{4-11} 
 &  &  & \textbf{6} & \textbf{0.8} & \textbf{0.2} & \textbf{{[}4, 4, 2, 2, 8, 2{]}} & \textbf{{[}1, 0, 1, 1, 1{]}} & \textbf{1.0} & \textbf{0.455} & \textbf{0.709} \\ \cline{4-11} 
  &  &  & {6} & {0.7} & {0.3} & {{[} 4, 4, 2, 2, 6, 2, 2{]}} & {{[} 1, 0, 1, 1, 1, 1 {]}} & {0.969} & {0.438 } & { 0.708} \\ \hline
\multirow{6}{*}{BiLSTM+RL+Fill} & \multirow{6}{*}{/} & \multirow{6}{*}{2} & 2 & 0.8 & 0.2 & {[}8, 12, 2{]} & {[}0, 1{]} & 0.938 & 0.455 & 0.727 \\ \cline{4-11} 
 &  &  & 4 & 0.8 & 0.2 & {[}2, 2, 4, 2, 6, 4, 2{]} & {[}0, 1, 0, 1, 1, 1{]} & 0.969 & 0.388 & 0.670 \\ \cline{4-11} 
 &  &  & 4 & 0.9 & 0.1 & {[}2, 2, 12, 4, 2{]} & {[}0, 1, 1, 1{]} & 1.0 & 0.504 & 0.738 \\ \cline{4-11} 
  &  &  & {6} & {0.9} & {0.1} & {{[}4, 4, 2, 2, 8, 2{]}} & {{[} 1, 1, 1, 1, 1{]}} & { 1.0} & {0.488 } & { 0.729} \\ \cline{4-11} 
 &  &  & \textbf{6} & \textbf{0.8} & \textbf{0.2} & \textbf{{[}8, 2, 2, 8, 2{]}} & \textbf{{[}0, 1, 1, 1{]}} & \textbf{1.0} & \textbf{0.471} & \textbf{0.719} \\  \cline{4-11} 
  &  &  & {6} & {0.7} & {0.3} & {{[} 4, 4, 2, 2, 8, 2 {]}} & {{[} 1, 0, 1, 1, 0 {]}} & { 0.938} & { 0.455} & {0.727 } \\ \hline
\multirow{6}{*}{\begin{tabular}[c]{@{}l@{}}LSTM+RL+\\ Dynamic-fill\end{tabular}} & \multirow{6}{*}{/} & \multirow{6}{*}{2} & 
{grades: 4} & {0.9} & {0.1} & {{[}2, 6, 2, 2, 8, 2{]}} & {{[}3, 1, 2, 2, 3 {]}} & { 1.0} & {0.558} & {0.763} \\ \cline{4-11} 
  &  &  & {grades: 4} & {0.8} & {0.2} & {{[}2, 2, 14, 4{]}} & {{[}0, 2, 2{]}} & {1.0} & {0.558} & {0.763} \\ \cline{4-11} 
 &  &  & \textbf{grades: 4} & \textbf{0.75} & \textbf{0.25} & \textbf{{[}2, 2, 4, 2, 2, 6, 4{]}} & \textbf{{[}1, 3, 0, 2, 3, 2{]}} & \textbf{1.0} & \textbf{0.43} & \textbf{0.692} \\ \cline{4-11} 
  &  &  & \begin{tabular}[c]{@{}l@{}}{grades: 4}\end{tabular} & {0.7} & {0.3} & {{[} 2, 6, 2, 8, 4 {]}} & {{[} 0, 1, 2, 2{]}} & { 0.938} & { 0.442 } & {0.720 } \\ \cline{4-11} 
   &  &  & \begin{tabular}[c]{@{}l@{}}{grades: 6}\end{tabular} & {0.8} & {0.2} & {{[}8, 2, 2, 10 {]}} & {{[}5, 2, 3 {]}} &  {1.0} & {0.521}  & {0.746}  \\ \cline{4-11} 
 &  &  & \begin{tabular}[c]{@{}l@{}}{grades: 6}\end{tabular} & 0.75 & 0.25 & {[}2, 2, 4, 2, 8, 4{]} & {[}4, 5, 0, 3, 4{]} & 0.969 & 0.397 & 0.677 \\ \bottomrule
\end{tabular}\\
\footnotesize{$^\dagger$ Sparsity of original matrix: 0.868.}\\
\end{table*}





\begin{table}[]
\centering
\caption{Complexity comparison of different methods.}
\label{table:R5828-complexity}
\begin{tabular}{l|l|l|l|l|l|l}
\toprule
Methods$^\dagger$ & \begin{tabular}[c]{@{}l@{}}Grid\\ size\end{tabular} & T & I & H & K & Complexity \\ \hline
LSTM+RL & 2 & 12 & 1 & 10 & 1 & \begin{tabular}[c]{@{}l@{}}$\mathcal{O}$(T(4IH+4H\textasciicircum{}2\\ +3H+HK))\end{tabular} \\ \midrule
\multirow{4}{*}{\begin{tabular}[c]{@{}l@{}}LSTM+RL\\ +Fill\end{tabular}} & \multirow{4}{*}{2} & \multirow{4}{*}{36} & 1 & 10 & 1 & \multirow{4}{*}{\begin{tabular}[c]{@{}l@{}}$\mathcal{O}$(T(4IH+4H\textasciicircum{}2\\ +3H+HK))\end{tabular}} \\ \cline{4-6}
 &  &  & 1 & 10 & 1 &  \\ \cline{4-6}
 &  &  & 1 & 10 & 1 &  \\ \cline{4-6}
 &  &  & 1 & 10 & 1 &  \\ \hline
\multirow{4}{*}{\begin{tabular}[c]{@{}l@{}}BiLSTM+RL\\ +Fill\end{tabular}} & \multirow{4}{*}{2} & \multirow{4}{*}{36} & 1 & 10 & 1 & \multirow{4}{*}{\begin{tabular}[c]{@{}l@{}}$\mathcal{O}$(2T(4IH+4H\textasciicircum{}2\\ +3H+HK))\end{tabular}} \\ \cline{4-6}
 &  &  & 1 & 10 & 1 &  \\ \cline{4-6}
 &  &  & 1 & 10 & 1 &  \\ \cline{4-6}
 &  &  & 1 & 10 & 1 &  \\ \hline
\multirow{3}{*}{\begin{tabular}[c]{@{}l@{}}LSTM+RL+\\ Dynamic-fill\end{tabular}} & \multirow{3}{*}{2} & \multirow{3}{*}{36} & 1 & 10 & 1 & \multirow{3}{*}{\begin{tabular}[c]{@{}l@{}}$\mathcal{O}$(T(4IH+4H\textasciicircum{}2\\ +3H+HK))\end{tabular}} \\ \cline{4-6}
 &  &  & 1 & 10 & 1 &  \\ \cline{4-6}
 &  &  & 1 & 10 & 1 &  \\ \bottomrule
\end{tabular}\\
\footnotesize{$^\dagger$: Taking \textit{QM7-5828} as example, T is the time step number of LSTM, I is the input size, H is the hidden size, K is the cell number. }\\
\end{table}

\begin{figure}[!ht]
\centering
\includegraphics[width=8.5cm]{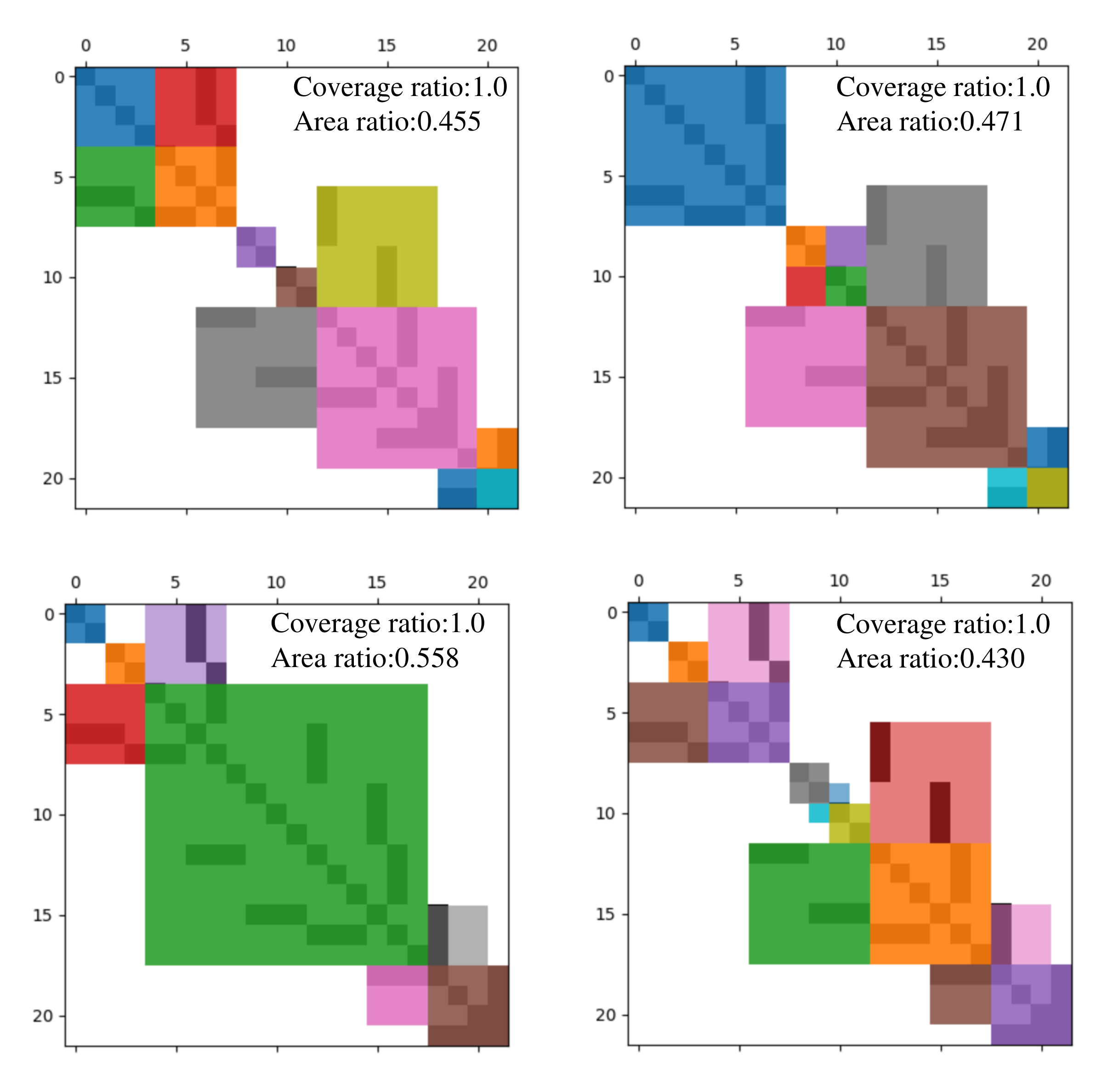}
\caption{Visualization of four representative mapping schemes {of \textit{QM7-5828}}, which are outstanding solutions (bold) in the Table.II. For a $22 \times 22$ matrix, it is burdensome and challenging to manually observe and output an optimal full-coverage scheme, but our agent can easily generate reasonable schemes with minimal area cost.}
\label{fig:R5828_visual}
\end{figure}

\begin{figure}[!ht]
\centering
\includegraphics[width=8.5cm]{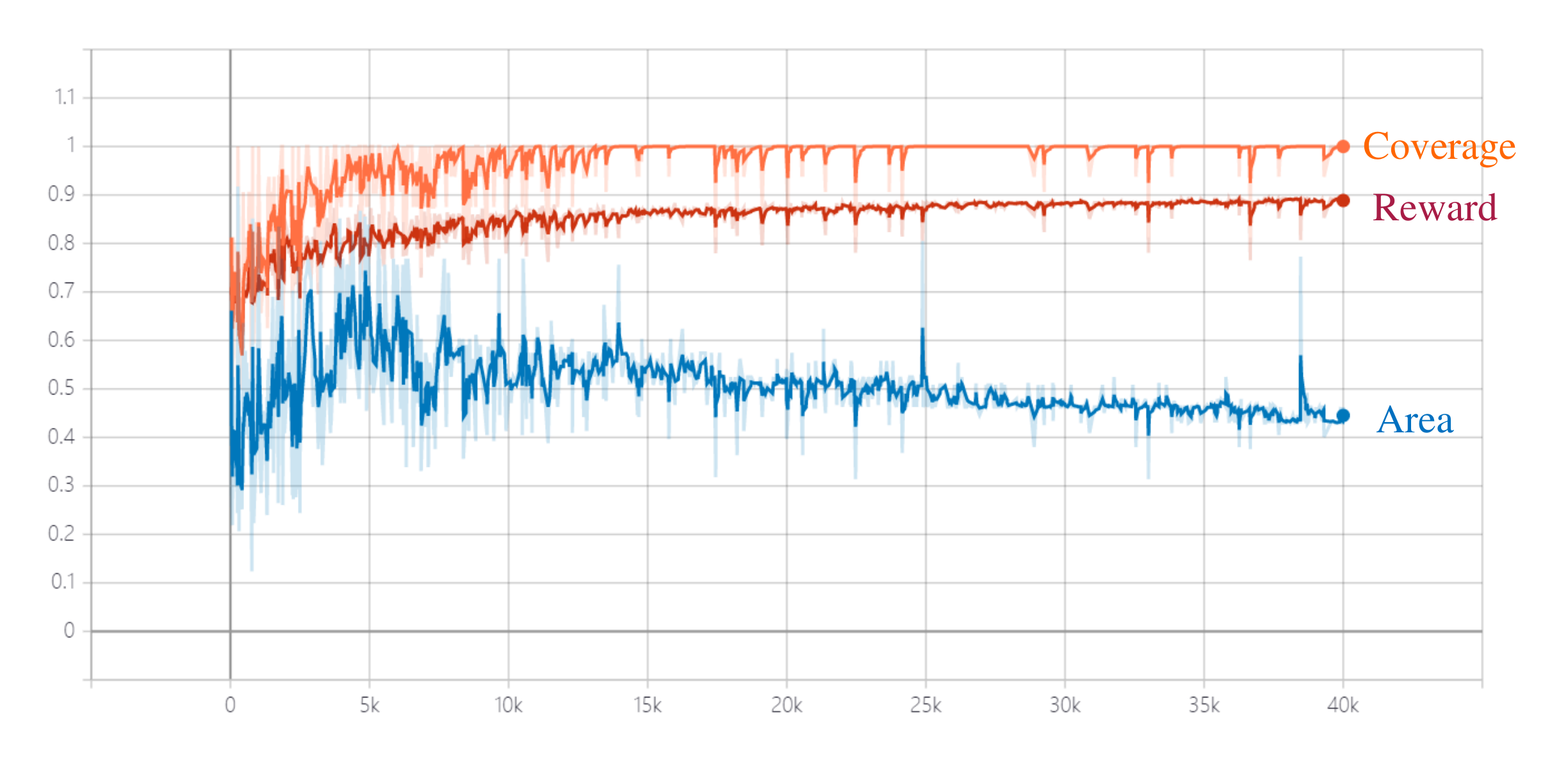}
\caption{The {optimization objective curves} of the coverage ratio {(as Eq.\eqref{eq:c_ratio})}, area ratio {(as Eq.\eqref{eq:a_ratio})}, and reward {of reinforcement learning (as Eq.\eqref{eq:multi-objective})} during the training on \textit{QM7-5828} graph data. After $5K$ epochs, the coverage ratio converges to 1 (with small fluctuation), and the area ratio converges to a specific smaller value compared with the early stage of the training.}
\label{fig:5828_curve}
\end{figure}

\begin{table*}[]
\caption{Experimental comparison results on large-scale matrix dataset: qh882 and qh1484.}
\begin{tabular}{c|l|l|l|l|l|l|l|l|l}
\toprule
\multirow{2}{*}{Dataset} & \multicolumn{1}{c|}{\multirow{2}{*}{\begin{tabular}[c]{@{}c@{}}Grid\\ size\end{tabular}}} & \multicolumn{1}{c|}{\multirow{2}{*}{\begin{tabular}[c]{@{}c@{}}Fill\\ grades\end{tabular}}} & \multicolumn{2}{l|}{Reward   ratio} & \multicolumn{2}{c|}{Solutions} & \multicolumn{1}{c|}{\multirow{2}{*}{\begin{tabular}[c]{@{}c@{}}Coverage\\ ratio\end{tabular}}} & \multicolumn{1}{c|}{\multirow{2}{*}{\begin{tabular}[c]{@{}c@{}}Area\\ ratio\end{tabular}}} & \multirow{2}{*}{Sparsity$^\dagger$} \\ \cline{4-7}
 & \multicolumn{1}{c|}{} & \multicolumn{1}{c|}{} & \multicolumn{1}{l|}{a} & 1-a & \multicolumn{1}{l|}{Diagonal-blocks size} & Fill-blocks size & \multicolumn{1}{c|}{} & \multicolumn{1}{c|}{} &  \\ \hline
\multirow{4}{*}{qh882} & 32 & 4 & \multicolumn{1}{l|}{0.7} & 0.3 & \multicolumn{1}{l|}{\begin{tabular}[c]{@{}l@{}}{[}32, 32, 32, 192, 96, 96, 64, 64, 96, \\ 96, 64, 18\end{tabular}} & \begin{tabular}[c]{@{}l@{}}{[}2, 3, 1, 2, 2, 2, 2, \\ 2, 2, 2, 2{]}\end{tabular} & 0.998 & 0.196 & 0.978 \\ \cline{2-10} 
 & 32 & 4 & \multicolumn{1}{l|}{0.8} & 0.2 & \multicolumn{1}{l|}{\begin{tabular}[c]{@{}l@{}}{[}32, 128, 96, 128, 96, 64, 64, 96, 96, \\ 32, 50{]}\end{tabular}} & \begin{tabular}[c]{@{}l@{}}{[}2, 2, 2, 2, 2, 2, 2, \\ 2, 3, 2{]}\end{tabular} & 0.998 & 0.204 & 0.979 \\ \cline{2-10} 
 & 32 & 6 & \multicolumn{1}{l|}{0.7} & 0.3 & \multicolumn{1}{l|}{\begin{tabular}[c]{@{}l@{}}{[}32, 32, 160, 160, 128, 96, 96, 128, \\ 32, 18{]}\end{tabular}} & \begin{tabular}[c]{@{}l@{}}{[}4, 4, 2, 3, 3, 3, 2,\\  5, 3{]}\end{tabular} & 0.995 & 0.2 & 0.979 \\ \cline{2-10} 
 & \textbf{32} & \textbf{6} & \multicolumn{1}{l|}{\textbf{0.8}} & \textbf{0.2} & \multicolumn{1}{l|}{\textbf{{[}32, 192, 160, 96, 160, 96, 64, 82{]}}} & \textbf{{[}2, 3, 4, 2, 3, 4, 3{]}} & \textbf{1.0} & \textbf{0.225} & \textbf{0.955} \\ \midrule
\multicolumn{1}{l|}{\multirow{4}{*}{qh1484}} & 32 & 4 & \multicolumn{1}{l|}{0.7} & 0.3 & \multicolumn{1}{l|}{\begin{tabular}[c]{@{}l@{}}{[}96, 32, 32, 288, 192, 160, 64, 32, 64, \\ 64, 32, 64, 32, 128, 32, 32, 64, 32, 44{]}\end{tabular}} & \begin{tabular}[c]{@{}l@{}}{[}1, 2, 1, 2, 2, 3, 3, 2, 2,\\  3,2, 3, 1, 2, 3, 2, 2, 3{]}\end{tabular} & 0.992 & 0.148 & 0.981 \\ \cline{2-10} 
\multicolumn{1}{l|}{} & 32 & 4 & \multicolumn{1}{l|}{0.8} & 0.2 & \multicolumn{1}{l|}{\begin{tabular}[c]{@{}l@{}}{[}96, 64, 288, 192, 128, 96, 128, 32, \\ 96, 32, 128, 64, 32, 96, 12{]}\end{tabular}} & \begin{tabular}[c]{@{}l@{}}{[}2, 2, 2, 2, 3, 2, 3, 2, \\ 3, 1, 3, 3, 2, 2{]}\end{tabular} & 0.999 & 0.185 & 0.985 \\ \cline{2-10} 
\multicolumn{1}{l|}{} & 32 & 6 & \multicolumn{1}{l|}{0.7} & 0.3 & \multicolumn{1}{l|}{\begin{tabular}[c]{@{}l@{}}{[}128, 224, 288, 224, 160, 64, 64, 160, \\ 64, 32, 64, 12{]}\end{tabular}} & \begin{tabular}[c]{@{}l@{}}{[}5, 2, 2, 3, 3, 2, 2, 4, 0, \\ 2, 4{]}\end{tabular} & 0.993 & 0.173 & 0.984 \\ \cline{2-10} 
\multicolumn{1}{l|}{} & \textbf{32} & \textbf{6} & \multicolumn{1}{l|}{\textbf{0.8}} & \textbf{0.2} & \multicolumn{1}{l|}{\textbf{\begin{tabular}[c]{@{}l@{}}{[}32, 96, 256, 288, 128, 96, 64, 64, \\ 128, 160, 64, 32, 64, 12{]}\end{tabular}}} & \textbf{\begin{tabular}[c]{@{}l@{}}{[}4, 5, 2, 4, 4, 4, 5, 3,\\  2, 2, 3, 3, 3{]}\end{tabular}} & \textbf{1.0} & \textbf{0.171} & \textbf{0.984} \\ \bottomrule
\end{tabular}\\
\label{table:882}
\footnotesize{$^\dagger$ Sparsity of \textit{qh882} is 0.995, sparsity of \textit{qh1484} is 0.997.}\\
\end{table*}

\begin{figure}[!ht]
\centering
\includegraphics[width=8.5cm]{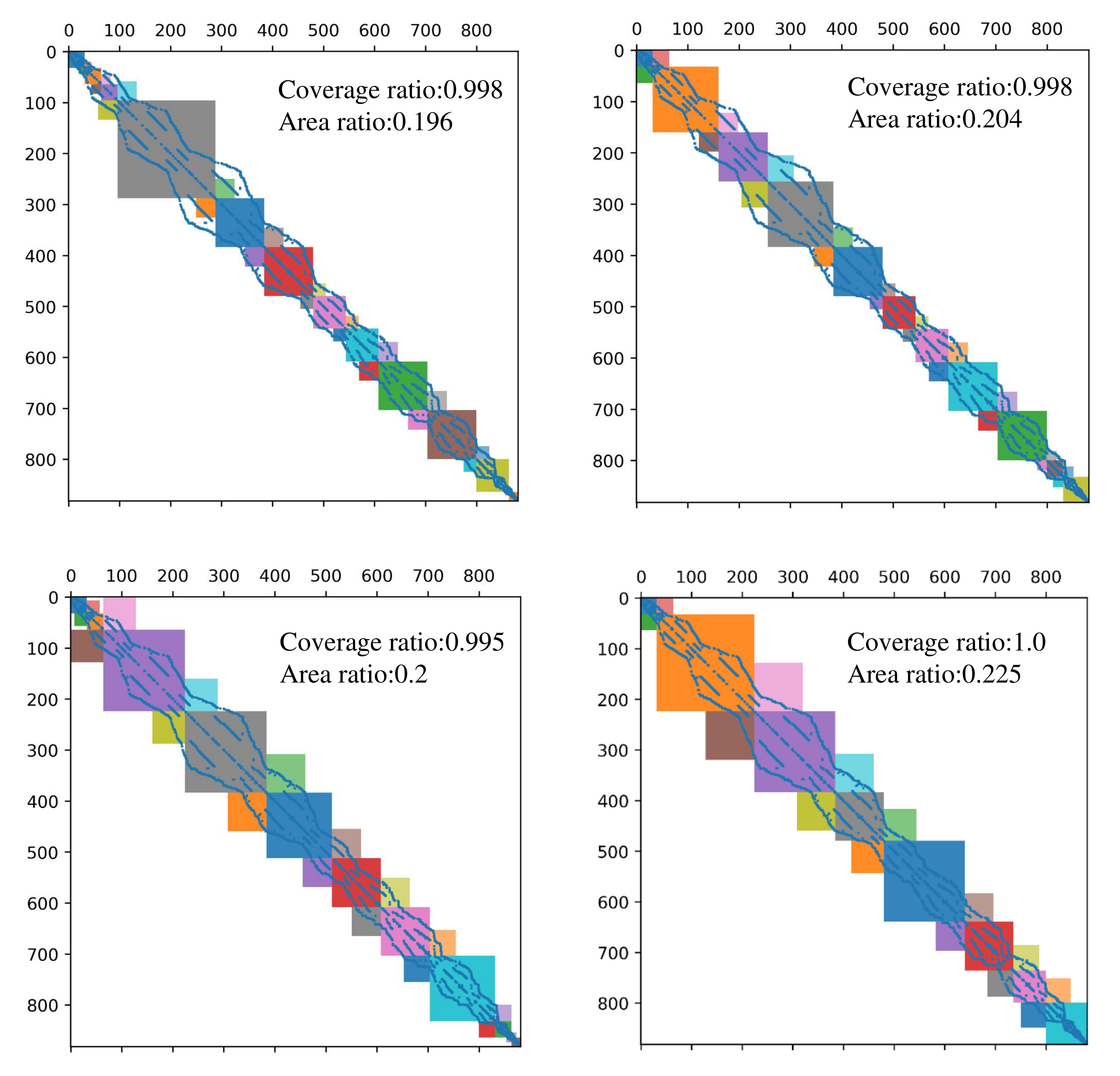}
\caption{Visualization of four representative mapping schemes of \textit{qh882} in Table.\ref{table:882}.}
\label{fig:882_visual}
\end{figure}

\begin{figure}[!ht]
\centering
\includegraphics[width=8.5cm]{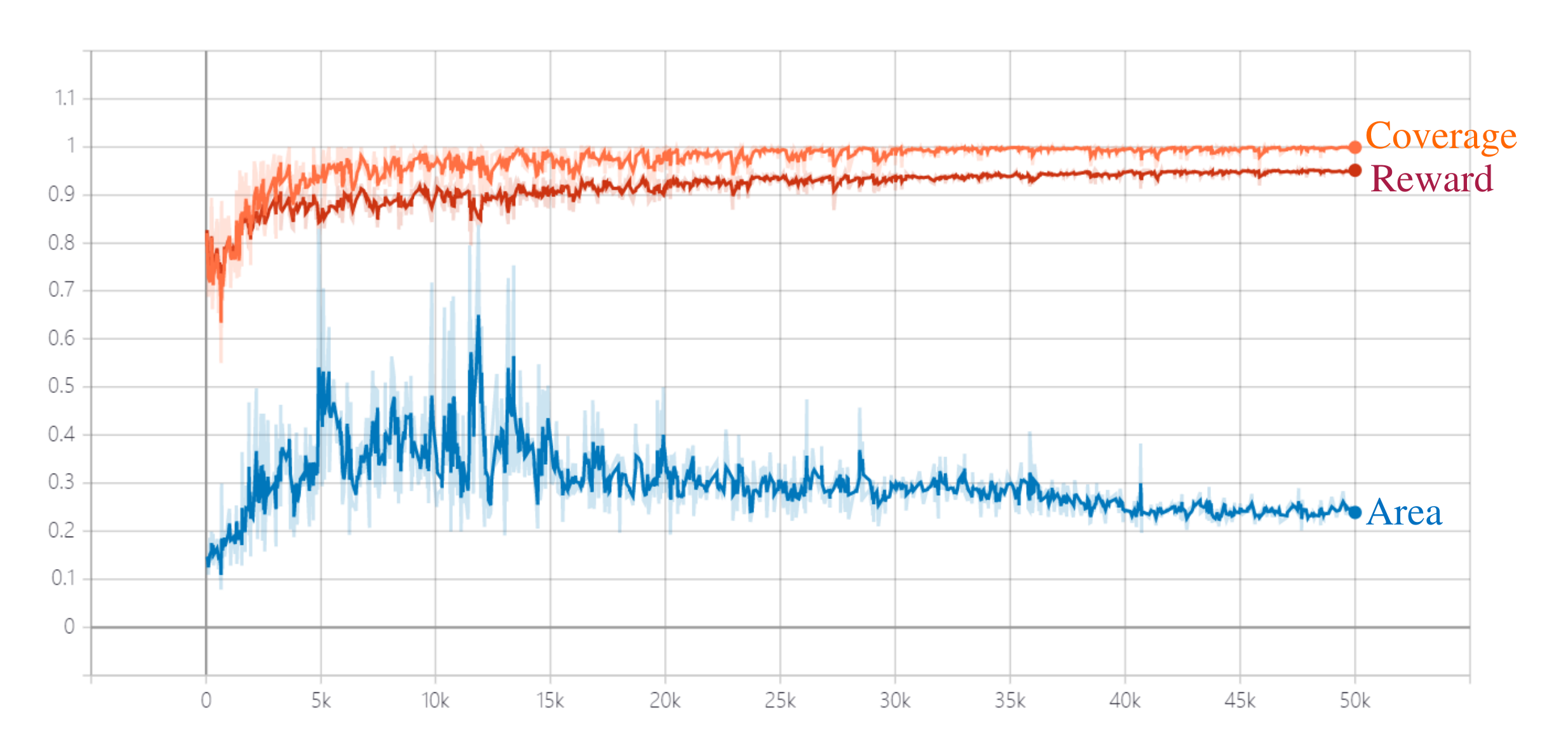}
\caption{The {optimization objective curves} of the coverage ratio {(as Eq.\eqref{eq:c_ratio})}, area ratio {(as Eq.\eqref{eq:a_ratio})}, and reward {of reinforcement learning (as Eq.\eqref{eq:multi-objective})} of the training on \textit{qh882} dataset. The optimization effect is consistent with that of \textit{QM7-5828} in Fig.\ref{fig:5828_curve}.}
\label{fig:882_curve}
\end{figure}

\begin{figure}[!ht]
\centering
\includegraphics[width=8.5cm]{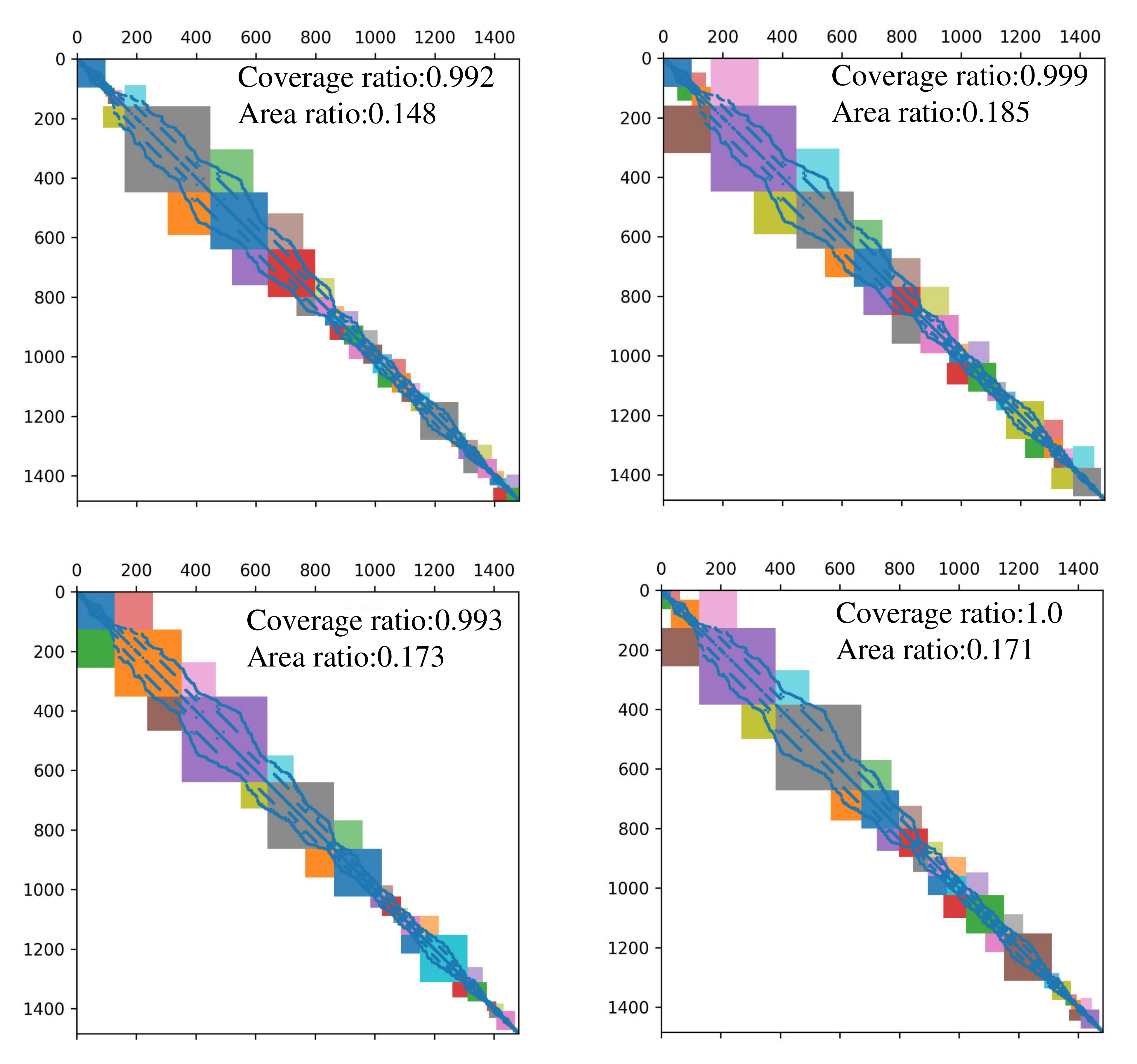}
\caption{The visualization of four representative mapping schemes on the \textit{qh1484} graph dataset in Table.\ref{table:882}.}
\label{fig:1484_visual}
\end{figure}

\begin{figure}[!ht]
\centering
\includegraphics[width=8.5cm]{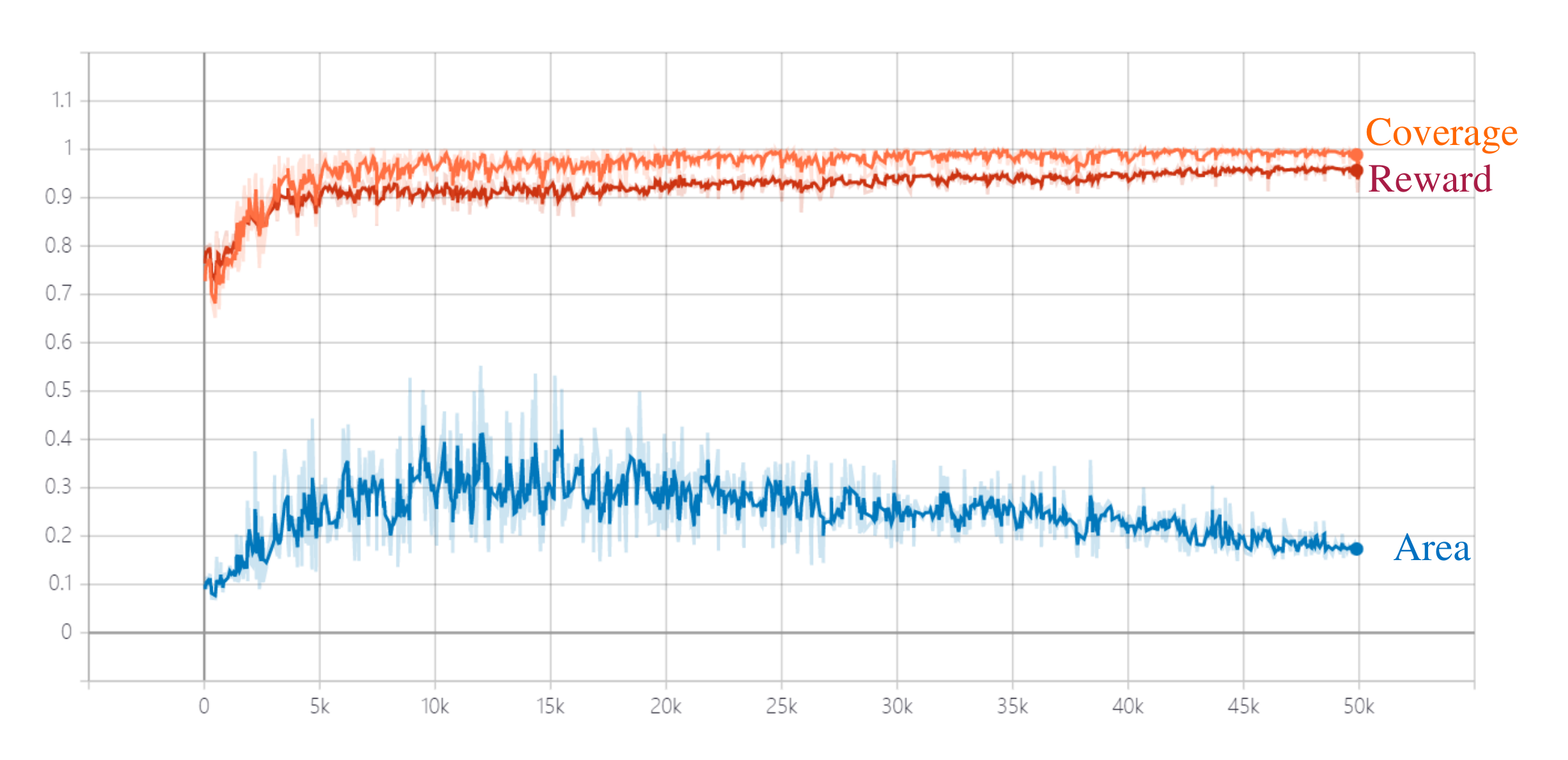}
\caption{The {optimization objective curves} of the coverage ratio {(as Eq.\eqref{eq:c_ratio})}, area ratio {(as Eq.\eqref{eq:a_ratio})}, and reward {(as Eq.\eqref{eq:multi-objective})} of the training on \textit{qh1484} graph data. The effect is consistent with that of \textit{qh882} in Fig.\ref{fig:882_curve}.}
\label{fig:1484_curve}
\end{figure}

Experimental results on \textit{QM7-5828} are presented in Table.\ref{table:R5828}, which showcases the comparison and ablation study. The comparison includes the fixed-size diagonal-block partition ({Vanilla}), fixed-size diagonal-block with additional fill-blocks ({Vanilla with fill}), our proposed method ``\textit{LSTM + RL}'' that only target the diagonal scheduling, while ``\textit{LSTM + RL + Fill}'' additionally utilize two series of blocks to ``fill the gap''. Based on the knowledge fusion among decision points, we study the effect of {BiLSTM}. Experiments show that, compared with {LSTM}, {BiLSTM} achieves no significant improvements. This also happens when we significantly increase the layer number of {LSTM} and the hidden size. {In Table.\ref{table:R5828-complexity}, we present the computational complexity comparison of different methods in Table.\ref{table:R5828}. The training of the LSTM (BPTT algorithm) costs most of the computational consumption, thus we omit the complexity of back-propagation of FCs.}
Visualizations of several typical outstanding mapping schemes (complete coverage scheme with different area ratios) are shown in Fig. \ref{fig:R5828_visual}. These four results are the promising solutions (in bold) in Table. \ref{table:R5828} in order.
{The objectives (reward) optimizing curves} of experiments on \textit{QM7-5828} are shown in Fig. \ref{fig:5828_curve}. With the continuous rise of the coverage ratio, the area ratio gradually converges to a small value after the big fluctuation in the early stage. The area ratio converges to an ideal ratio, but not at the expense of the coverage ratio, which can steadily converge to 1 after 40k epochs of training under an Intel CPU (without GPU). {In the later stage of our experiment, the controller is basically convergent, and the slight sampling fluctuation of the objectives is normal, because our model is a probabilistic sampling model, even if the LSTM controller parameters are fully converged, the sampling is still probabilistic.}

Further experiments based on large-scale graph data/matrix (\textit{qh882}, \textit{qh1484}) are shown in Table. \ref{table:882}, in which we directly adopt \textit{``LSTM + RL + Dynamic''} strategy.
To reduce the scale of the problem, we partition the original matrix into grids. Considering the grid size $k \times k$, thus the grid number is $N = D/k$, where $D$ is the size of the matrix.
Empirically, we set the grid size of \textit{qh882} and \textit{qh1484} to be 32, and the fill grades number to be 4, the action space is $2 ^{\lfloor 882/32 \rfloor} \times 4 ^ {\lfloor 882/32 \rfloor}  = 2.4 \times 10 ^ {24}$, and $2^ {\lfloor  1484/32 \rfloor} \times  4 ^ {\lfloor 1484/32 \rfloor} = 3.4 \times 10^{41}$, respectively, following the calculation rule of action space in Table. \ref{table:notation}.
As shown in Table. \ref{table:882}, fill-block size grades number of $6$ can achieve better results (\textit{qh882}: coverage ratio 1 with area ratio 0.225, \textit{qh1484}: Coverage ratio 1 with area ratio 0.171) than that of $4$. That means more fine-grained fill size grades can achieve better results, this is also expected because the utilization of the mapped blocks is high (with lower sparsity) in this way. Visualizations of \textit{qh882} mapping schemes are shown in Fig. \ref{fig:882_visual}, and {the objectives optimizing curves} of the promising configuration (bold in Table.\ref{table:882}) are shown in Fig. \ref{fig:882_curve}. Similarly, the visualization of \textit{qh1484} mapping scheme is shown in Fig. \ref{fig:1484_visual}, and {the corresponding objectives optimizing curves} are shown in Fig. \ref{fig:1484_curve}.

In real-world scenarios, the grid size is set subject to the allowable crossbar's size. The complexity of the peripheral circuit also depends on the grid size and the granularity of the dynamic-fill grade number, which ultimately jointly affect the area consumption of the final complete coverage solution. In principle, a smaller grid size and much more grades of the dynamic-fill-blocks will result in better mapping performance (complete coverage solutions with less area cost). Unavoidably, this also increases the action space of reinforcement learning, namely, the difficulty of agent optimization. These hyper-parameters can be tuned according to actual deployment scenarios, which demonstrates that our framework is flexible and scalable.

\section{Conclusion}
Some recent works have made attempts at the efficient computation of sparse graphs, e.g., mapping the large-scale graph data by block coverage around the diagonal after matrix reordering. But their mapping methods are fixed or static, instead of proposing an effective mapping schedule scheme, which should be feasible, scalable, and flexible.
Based on the background of PIM crossbar computing and the characteristics of block coverage schedule problems, we put forward the principles of the mapping framework as the criterion of subsequent research.
We creatively propose to formulate this problem into a sequential decision-making problem whose solution space is equal to the 0-1 integer programming problem, and creatively propose the dynamic-fill method to ``fill the gaps'', which well meets our proposed basic principles of the mapping framework.
We model the sequential decision-making problem using a generation agent consisting of LSTM and FCs, and leverage the reinforcement learning algorithm (REINFORCE) to optimize it.
Finally, the comparison and ablation experiments on a small dataset and two large matrix data show that flexible and scalable mapping schemes can be generated with limited training epochs and time cost, and are suitable for the deployment and compilation systems. {This method may also be extended to other PIM architectures, not limited to the memristive device-based platforms. Our proposed basic principles of mapping large-scale sparse graphs on memristive crossbars are worthy to be spread to the research community.}

From our perspective, the limitation lies in that we only exploit the fundamental sequence-to-sequence model. Although we have tried to increase LSTM's model scale (layers and hidden size), we did not carry out in-depth discussions and research on the knowledge fusion model, which may reach better decision-making performance. Similarly, the fundamental policy gradient algorithm is utilized. {In addition, our method is based on matrix reordering, by which to reduce the matrix bandwidth. The matrix needs to be a symmetric square matrix, such as a graph adjacency matrix. Therefore, there exist some limitations in the generalization of the scheme. Currently, it is not suitable for popular image classification models, e.g., CNNs or Transformers.}
Future, we plan to study the fusion of the automatic mapping scheme and the sparse storage (CSC, CSR, COO), thus further accelerating the pipeline of the graph processing on memristive crossbars of PIM/CIM platforms. {Further, our proposed sparsity-aware mapping learning algorithm may be further extended with the combination of sparsity \cite{DBLP:conf/nvmsa/SongLWLC17} and fault-tolerant training, or some device-circuit non-idealities of memristive crossbars, e.g., variation and defect \cite{DBLP:conf/dac/LiuLCLWH15,DBLP:journals/fgcs/JinPW20,DBLP:conf/iccad/GaolZYLSZ21}.}

\section*{Acknowledgments}
This work is supported in part by the Zhejiang provincial ``Ten Thousand Talents Program'' (2021R52007), the Science and Technology Innovation 2030-Major Project (2021ZD0114300), the Joint Research Project of Zhejiang Lab (K2022DA0AM01), and the National Key R\&D Program of China (2022YFB4500405). 

\balance
\bibliographystyle{IEEEtran}
\bibliography{ref}

\end{document}